\documentclass[runningheads]{llncs}
\usepackage{graphicx}
\usepackage{amsmath,amssymb} 
\usepackage{color}

\usepackage[hidelinks]{hyperref}

\usepackage{subfig}

\usepackage{lipsum}

\usepackage[dvipsnames]{xcolor}
\usepackage{enumitem}
\usepackage[normalem]{ulem} 
\usepackage{framed}
\usepackage{tcolorbox}
\usepackage{wrapfig,lipsum}
\definecolor{EPIC-COLOR}{HTML}{ED323E}
\definecolor{EPIC-COLOR2}{HTML}{00D6D6}
\usepackage{multirow} 
\usepackage{ifthen} 
\usepackage{makecell} 
\definecolor{EPIC-BLUE}{HTML}{00B6D6} 
\definecolor{EPIC-RED}{HTML} {ED323E} 
\usepackage{multicol} 
\usepackage{booktabs} 

\newcommand{\EPICTitle}{\textcolor{EPIC-COLOR}{{EPIC-KITCHENS}}}

\newcommand{\fps}{\textit{fps}}

\DeclareMathOperator*{\tvl}{TV-L_1\!}

\newcommand{\EPIC}{\textcolor{EPIC-COLOR}{{EPIC-KITCHENS}}}

\begin{document}
\pagestyle{headings}
\mainmatter

\title{Scaling Egocentric Vision:\\ The \EPICTitle{} Dataset} 


\titlerunning{Scaling Egocentric Vision:\\ The \EPICTitle{} Dataset}

\authorrunning{D. Damen et al}

\author{Dima Damen$^1$ \and
Hazel Doughty$^1$ \and
Giovanni Maria Farinella$^2$ \and 
Sanja Fidler$^3$ \and
Antonino Furnari$^2$ \and 
Evangelos Kazakos$^1$ \and
Davide Moltisanti$^1$ \and \\
Jonathan Munro$^1$ \and
Toby Perrett$^1$ \and
Will Price$^1$ \and 
Michael Wray$^1$}



\institute{$^1$Uni. of Bristol, UK $\quad ^2$Uni. of Catania, Italy, $\quad ^3$Uni. of Toronto, Canada
}



\maketitle
\vspace{-2mm}
\begin{abstract}
First-person vision is gaining interest as it offers a unique viewpoint on people's interaction with objects, their attention, and even intention. However, progress in this challenging domain has been relatively slow due to the lack of sufficiently large datasets.
In this paper, we introduce \EPIC{}, a large-scale egocentric video benchmark recorded by 32 participants in their native kitchen environments. Our videos depict \textbf{non-scripted} daily activities: we simply asked each participant to start recording every time they entered their kitchen. Recording took place in 4 cities (in North America and Europe) by participants belonging to 10 different nationalities, resulting in highly diverse cooking styles. Our dataset features 55 hours of video consisting of 11.5M frames, which we densely labelled for a total of $39.6$K action segments and $454.3$K object bounding boxes. Our annotation is unique in that we had the participants narrate their own videos (after recording), thus reflecting true intention, and we crowd-sourced ground-truths based on these. 
We describe our object, action and 
anticipation challenges, and evaluate several baselines over two test splits,
\textit{seen} and \textit{unseen} kitchens.

\vspace*{-8pt}
\end{abstract}




\begin{figure}[t]
\centering
\includegraphics[width=1\columnwidth]{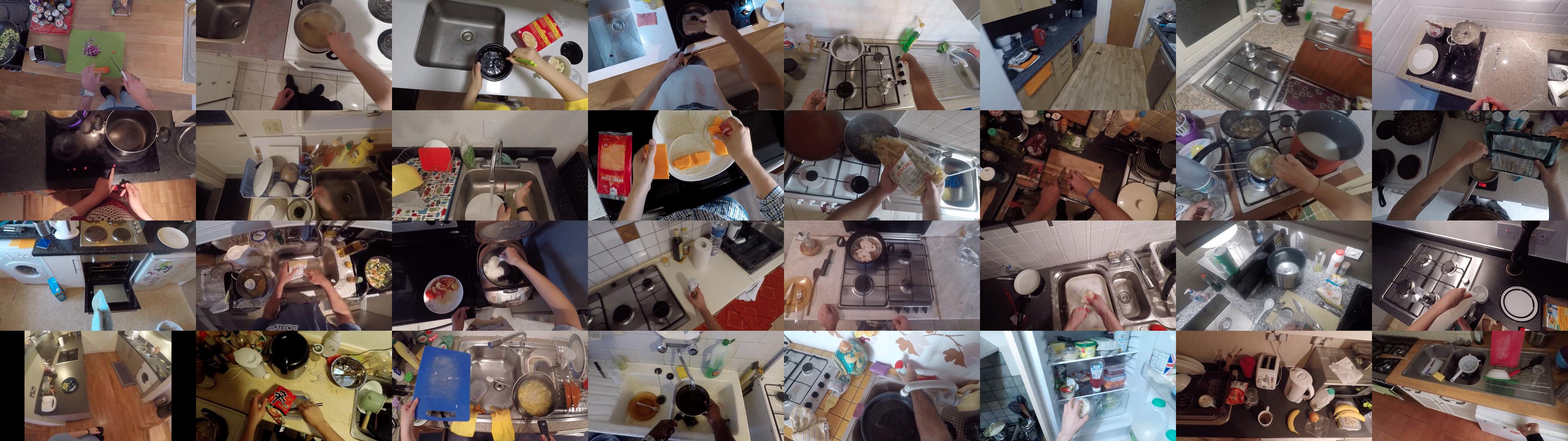}\\\vspace*{6pt}
\includegraphics[width=1\columnwidth]{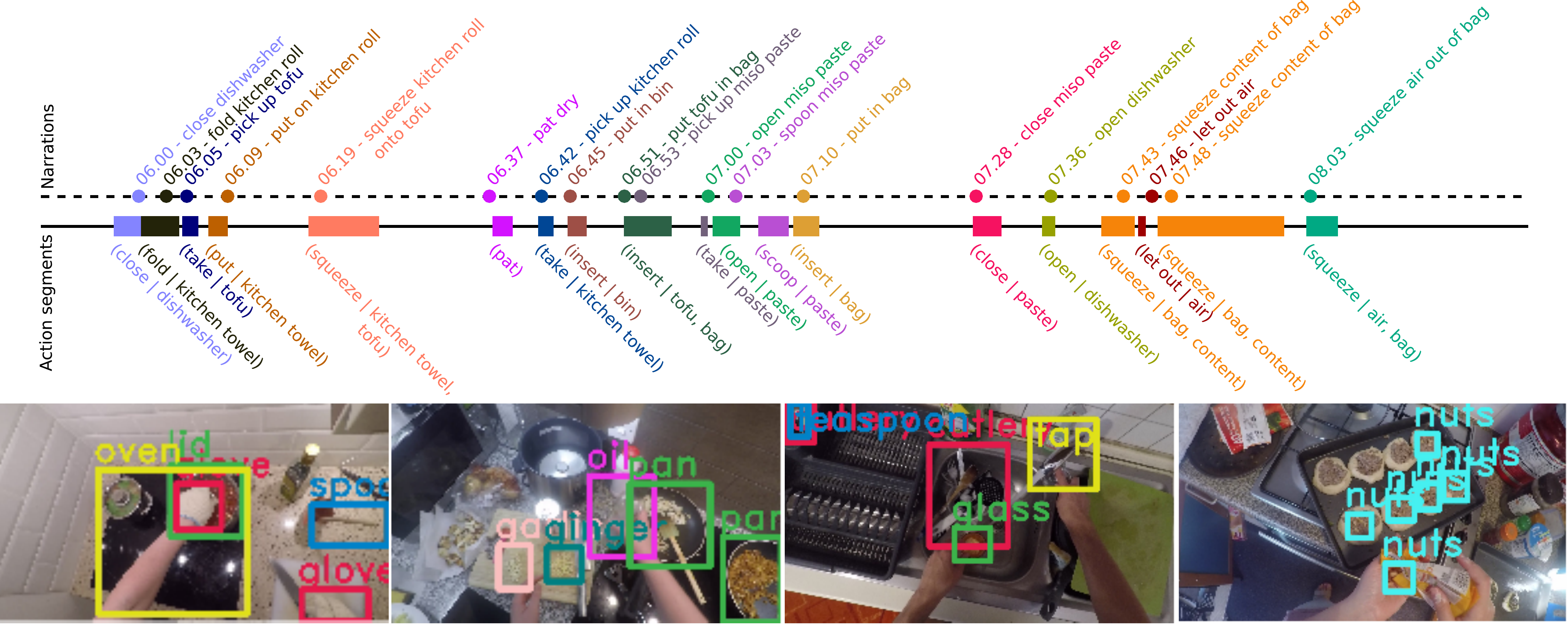}
\caption{From top: Frames from the 32 environments; Narrations by participants used to annotate action segments; Active object bounding box annotations}
\label{fig:epic-wall}
\end{figure}

%
\vspace*{-20pt}
\section{Introduction}\label{sec:introduction}


In recent years, we have seen significant progress in many domains such as image classification~\cite{resnet}, object detection~\cite{ren2015faster}, captioning~\cite{Karpathy14} and visual question-answering~\cite{VQA}. This success has in large part been due to advances in deep learning~\cite{alexnet} as well as the availability of large-scale image benchmarks~\cite{pascal,imagenet,coco,ade}. 
While gaining attention, work in video understanding has been more scarce, mainly due to the lack of annotated datasets. This has been changing recently, with the release of the action classification benchmarks such as~\cite{Goyal2017,Haija2016,zhao2017slac,moviedescription,movieqa,fouhey2017lifestyle}. With the exception of~\cite{movieqa}, most of these datasets contain videos that are very short in duration, i.e., only a few seconds long, focusing on a single action. Charades~\cite{Sigurdsson2016} makes a step towards activity recognition by collecting 10K videos of humans performing various tasks in their home. While this dataset is a nice attempt to collect daily actions, the videos have been recorded in a scripted way, by asking AMT workers to act out a script in front of the camera. This makes the videos look oftentimes less natural, and they also lack the progression and multi-tasking of actions that occur in real life. 

Here we focus on first-person vision, which offers a unique viewpoint on people's daily activities. This data is rich as it reflects our goals and motivation, ability to multi-task, and the many different ways to perform a variety of important, but mundane, everyday tasks (such as cleaning the dishes). 
Egocentric data has also recently been proven valuable for human-to-robot imitation learning~\cite{Nair2017,Zhang2018}, and has a direct impact on HCI applications. However, datasets to evaluate first-person vision algorithms~\cite{EGTEA,sigurdsson2018charadesego,Damen2014a,Fathi2012,Pirsiavash2012,de2008guide} have been significantly smaller in size than their third-person counterparts, often captured in a single environment~\cite{EGTEA,Damen2014a,Fathi2012,de2008guide}.
Daily interactions from wearable cameras are also scarcely available online, making this a largely unavailable source of information. 

In this paper, we introduce \EPIC{}, a  large-scale egocentric dataset. Our data was collected by 32 participants, belonging to 10 nationalities, in their native kitchens (Fig.~\ref{fig:epic-wall}). The participants were asked to capture all their daily kitchen activities, and record sequences regardless of their duration. The recordings, which include both video and sound, not only feature the typical interactions with one's own kitchenware and appliances, but importantly show the natural multi-tasking that one performs, like washing a few dishes amidst cooking. Such parallel-goal interactions have not been captured in existing datasets, making this both a more realistic as well as a more challenging set of recordings.
A video introduction to the recordings is available at: \textcolor{blue}{\underline{\url{http://youtu.be/Dj6Y3H0ubDw}}}.

\vspace*{6pt}
 
Altogether, \EPIC{} has 55hrs of recording, densely annotated with start/end times for each action/interaction, as well as bounding boxes around objects subject to interaction. We describe our object, action and anticipation challenges, and report baselines in two scenarios, i.e., \textit{seen} and \textit{unseen} kitchens. The dataset and leaderboards to track the community's progress on all challenges, with held out test ground-truth are at: \textcolor{blue}{\underline{\url{http://epic-kitchens.github.io}}}.

\begin{table}[t!]
\vspace{-2mm}
\begin{center}
\caption{Comparative overview of relevant datasets \tiny{$^*$action classes with $>50$ samples}}
\label{tab:datasets}
\resizebox{\columnwidth}{!}{%
\begin{tabular}{|l|c|c|c|c|r|c|r|c|r|c|c|c|c|}
\hline

 & &\textbf{Non-} &\textbf{Native} & &  & \textbf{Sequ-} &\textbf{Action} & \textbf{Action} & \textbf{Object} & \textbf{Object} 
& \textbf{Partici-} & \textbf{No.}\\
\textbf{Dataset} & \textbf{Ego?} &\textbf{Scripted?} &\textbf{Env?} &\textbf{Year} & \textbf{Frames} & \textbf{ences} & \textbf{Segments} &\textbf{Classes} & \textbf{BBs} & \textbf{Classes} 
& \textbf{pants} &\textbf{Env.s} \\
\hline
\hline
\EPIC{} &$\checkmark$ &$\checkmark$ &$\checkmark$ &2018 &11.5M  &432  &39,596 &149* &454,255 &323  &32 & 32 \\
\hline
\hline
EGTEA Gaze+~\cite{EGTEA}  &$\checkmark$ &$\times$ &$\times$ &2018 & 2.4M 
&86  &10,325  &106 &0 &0 &32 & 1 \\
Charades-ego~\cite{sigurdsson2018charadesego} &{\tiny 70\%} $\checkmark$ &$\times$ &$\checkmark$ &2018 &2.3M &2,751 & 30,516 &157 &0 &38 &71 &N/A \\
BEOID~\cite{Damen2014a} &$\checkmark$ &$\times$ &$\times$ &2014 &0.1M
&58  &742  &34 &0 &0 &5 & 1 \\
GTEA Gaze+~\cite{Fathi2012}  &$\checkmark$ &$\times$ &$\times$ &2012 & 0.4M
&35  &3,371  &42 &0 &0 &13 & 1 \\
ADL~\cite{Pirsiavash2012} &$\checkmark$ &$\times$ &$\checkmark$ &2012 &1.0M 
&20  &436  &32 &137,780 &42 &20 &20 \\
CMU~\cite{de2008guide} &$\checkmark$ &$\times$ &$\times$ &2009 & 0.2M
&16  &516 &31 &0 &0 &16 &1 \\
\hline\hline
YouCook2~\cite{Zhou2017}& $\times$ & $\checkmark$ & $\checkmark$ & 2018 & {\tiny @30fps} 15.8M & 2,000 & 13,829 & 89 & 0 & 0 & 2K & N/A \\
VLOG \cite{fouhey2017lifestyle} & $\times$ & $\checkmark$ & $\checkmark$ & 2017 & 37.2M & 114K & 0 & 0 & 0 & 0 & 10.7K & N/A\\
Charades \cite{Sigurdsson2016} &$\times$ &$\times$ &$\checkmark$ &2016 &7.4M & 9,848 &67,000  & 157 &0 &0 & N/A & 267\\
Breakfast \cite{Kuehne2014}  &$\times$ &$\checkmark$ &$\checkmark$ &2014 & 
3.0M
& 433 & 3078 & 50 &0 &0 & 52 & 18 \\
50 Salads \cite{Stein2013} &$\times$ &$\times$ &$\times$ &2013 & 0.6M
& 50 & 2967 & 52 &0 &0 & 25 & 1 \\
MPII Cooking 2 \cite{Rohrbach2012} &$\times$ &$\times$ &$\times$ &2012 & 2.9M
& 273 & 14,105 & 88 &0 &0 & 30 & 1 \\
\hline
\end{tabular}}
\end{center}
\vspace{-4mm}
\end{table}

\section{Related Datasets}
We compare \EPIC{} to four commonly-used~\cite{Damen2014a,Fathi2012,Pirsiavash2012,de2008guide} and two recent~\cite{EGTEA,sigurdsson2018charadesego} egocentric datasets in Table~\ref{tab:datasets}, as well as six 
third-person activity-recognition datasets \cite{fouhey2017lifestyle,Sigurdsson2016,Zhou2017,Kuehne2014,Stein2013,Rohrbach2012} that focus on object-interaction activities. 
We exclude  egocentric datasets that focus on inter-person interactions~\cite{Alletto2015,Fathi2012b,Ryoo2013}, as these target a different research question. 

A few datasets aim at capturing activities in native environments, most of which are recorded in third-person~\cite{Goyal2017,fouhey2017lifestyle,Sigurdsson2016,sigurdsson2018charadesego,Kuehne2014}. ~\cite{Kuehne2014} focuses on cooking dishes based on a list of breakfast recipes. In~\cite{fouhey2017lifestyle}, short segments linked to interactions with 30 daily objects are collected by querying YouTube, while~\cite{Goyal2017,Sigurdsson2016,sigurdsson2018charadesego} are scripted -- subjects are requested to enact a crowd-sourced storyline~\cite{Sigurdsson2016,sigurdsson2018charadesego}\footnote{\tiny{In discussion with the primary author and based on our analysis of the released footage, around 70\% of videos in Charades-ego are truly egocentric (i.e. recorded using a wearable camera with the action performed by the wearer). We use this percentage in reporting statistics on this dataset. }} or a given action~\cite{Goyal2017}, which oftentimes results in less natural looking actions.
All egocentric datasets similarly use scripted activities, i.e. people are told what actions to perform. When following  instructions, participants perform steps in a sequential order, as opposed to the more natural real-life scenarios addressed in our work, which involve multi-tasking, searching for an item, thinking what to do next, changing one's mind or even unexpected surprises. \EPIC{} is most closely related to the ADL dataset~\cite{Pirsiavash2012} which also provides egocentric recordings in native environments. However, our dataset is substantially larger: it has 11.5M frames vs 1M in ADL, 90x more annotated action segments, and 4x more object bounding boxes, making it the largest first-person dataset to date. 

\section{The \EPIC{} Dataset}
\label{sec:dataset}

In this section, we describe our data collection and annotation pipeline. We also present various statistics, showcasing different aspects of our collected data.

\vspace{-1mm}
\subsection{Data Collection}
\label{sec:recording}

\begin{figure}[t]
\centering
{\includegraphics[width=0.7\columnwidth]{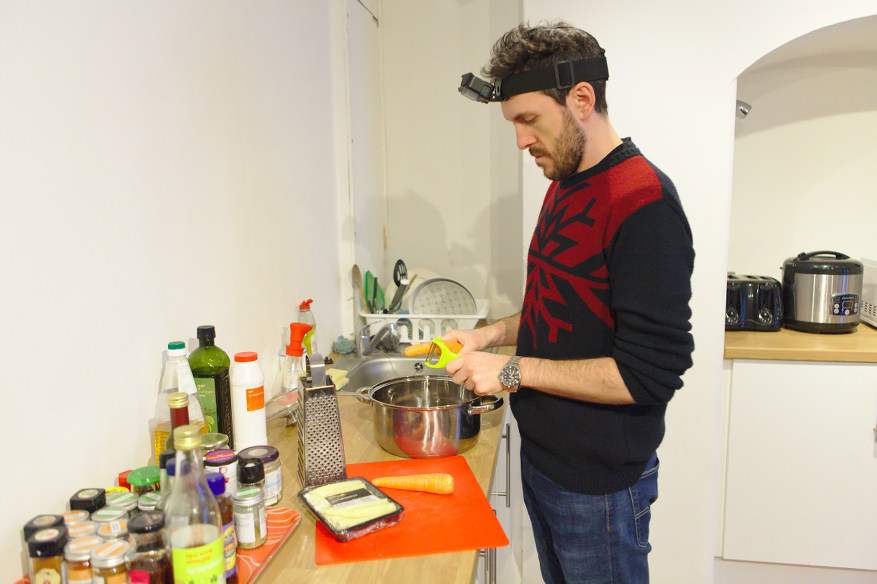}}
\caption{Head-mounted GoPro used in dataset recording}
\label{fig:setup}
\end{figure}
\begin{figure}[t!]
\begin{tcolorbox}[width=\linewidth,boxrule=1pt,colback=blue!4,left=2pt,right=2pt,top=2pt,bottom=2pt]
\scriptsize{
\noindent Use any word you prefer. Feel free to vary your words or stick to a few.\\[-2.3mm]

\noindent Use present tense verbs (e.g. cut/open/close).\\[-2.3mm]

\noindent Use verb-object pairs (e.g. “wash carrot”).\\[-2.3mm]

\noindent You may (if you prefer) skip articles and pronouns (e.g. ``cut kiwi'' rather than ``I cut the kiwi'').\\[-2.3mm] 

\noindent Use propositions when needed (e.g. ``pour water into kettle'').\\[-2.3mm]

\noindent Use `and' when actions are co-occurring (e.g. ``hold mug and pour water'').\\[-2.3mm]

\noindent If an action is taking long, you can narrate again (e.g. ``still stirring soup'').}
\end{tcolorbox}
\vspace{-3mm}
\caption{Instructions used to collect video narrations from our participants}
\label{fig:narrationInstruction}
\vspace{-2mm}
\end{figure}

The dataset was recorded by 32 individuals in 4 cities in different countries (in North America and Europe): 15 in Bristol/UK, 
8 in Toronto/Canada, 
8 in Catania/Italy 
and 1 in Seattle/USA
between May and Nov~2017. 
Participants were asked to capture all kitchen visits \textit{for three consecutive days}, with the recording starting immediately before entering the kitchen, and only stopped before leaving the kitchen. They recorded the dataset voluntarily and were not financially rewarded. The participants were asked to be in the kitchen alone for all the recordings, thus capturing only one-person activities. We also asked them to remove all items that would disclose their identity such as portraits or mirrors. 
Data was captured using a head-mounted GoPro with an adjustable mounting to control the viewpoint for different environments and participants' heights. Before each recording, the participants checked the battery life and viewpoint, using the GoPro Capture app, 
so that their stretched hands were approximately located at the middle of the camera frame. The camera was set to linear field of view, 59.94\fps{} and Full HD resolution of 1920x1080, however some subjects made minor changes like wide or ultra-wide FOV or resolution, as they recorded multiple sequences in their homes, and thus were switching the device off and on over several days. Specifically, 1\% of the videos were recorded at 1280x720 and 0.5\% at 1920x1440. Also, 1\% at 30\fps, 1\% at 48\fps{} and 0.2\% at 90\fps.

\begin{figure}[t]
\includegraphics[width=0.25\columnwidth]{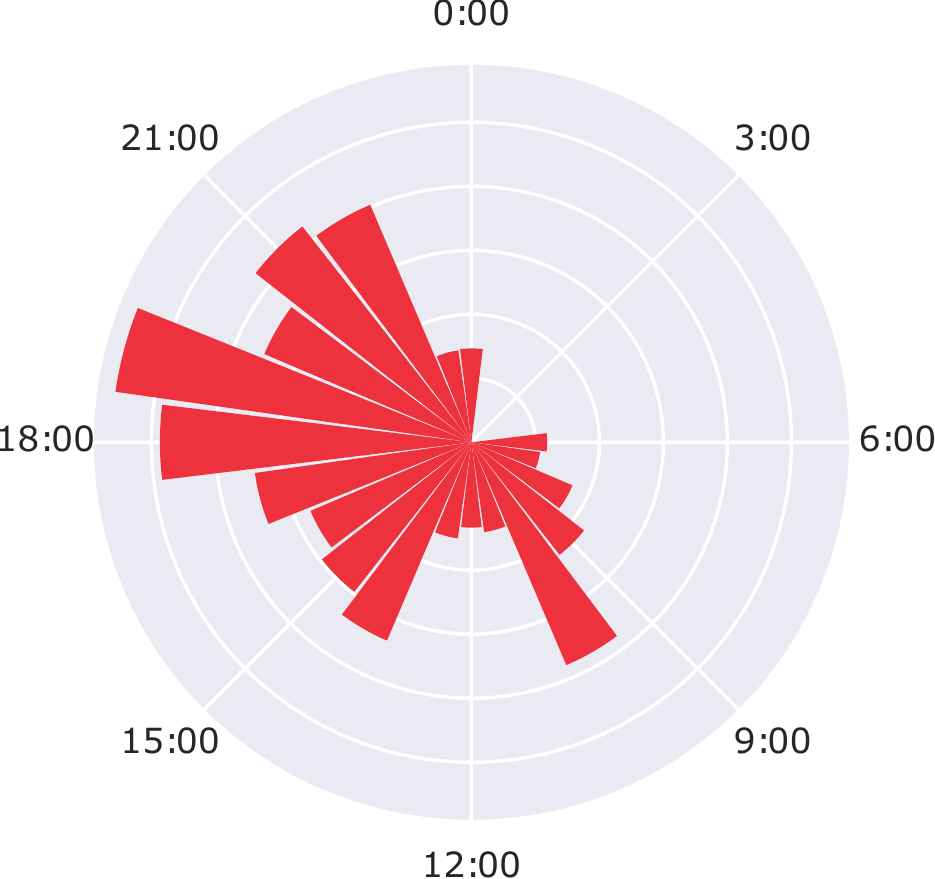}
\includegraphics[width=0.28\columnwidth]{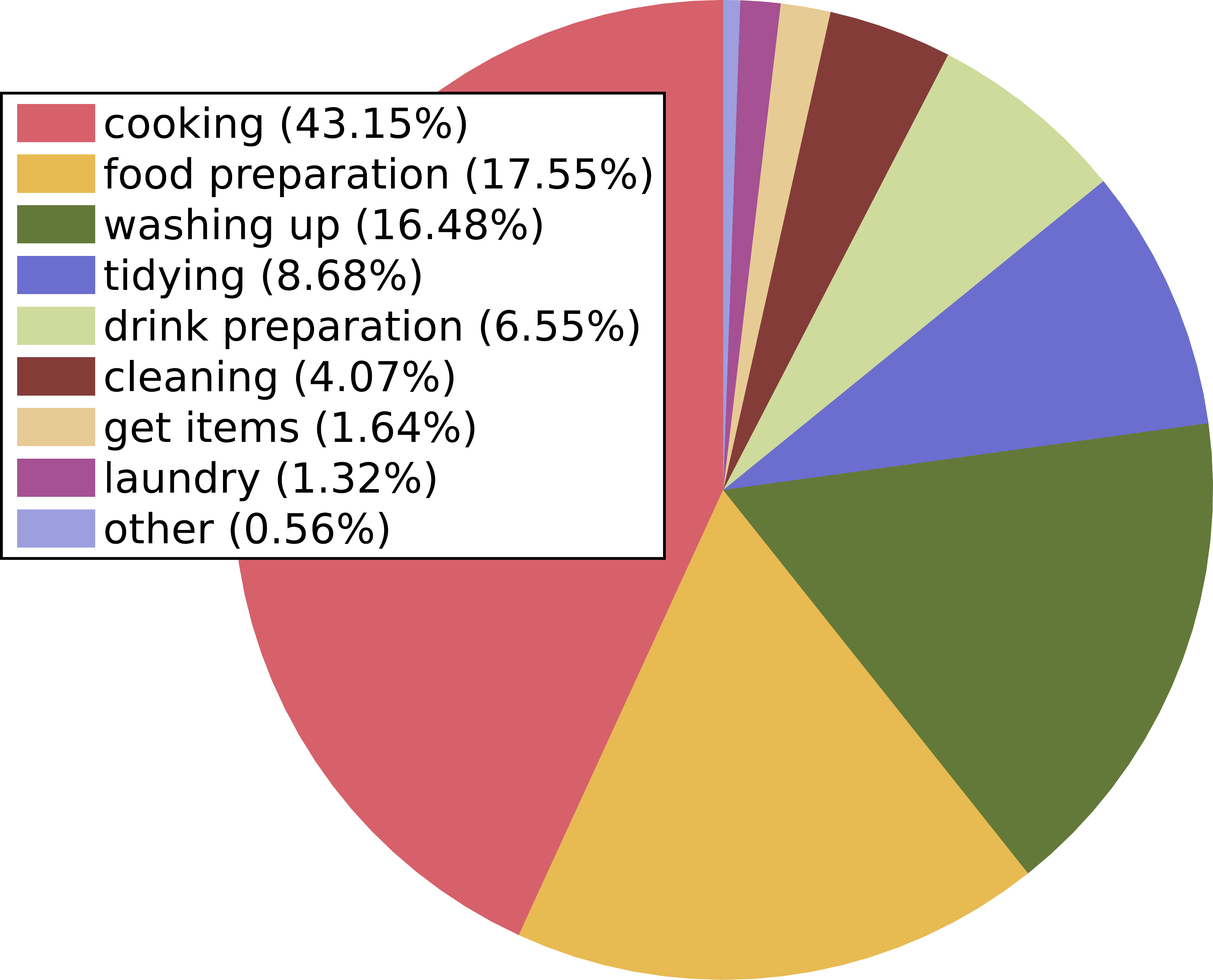}
\includegraphics[width=0.45\columnwidth]{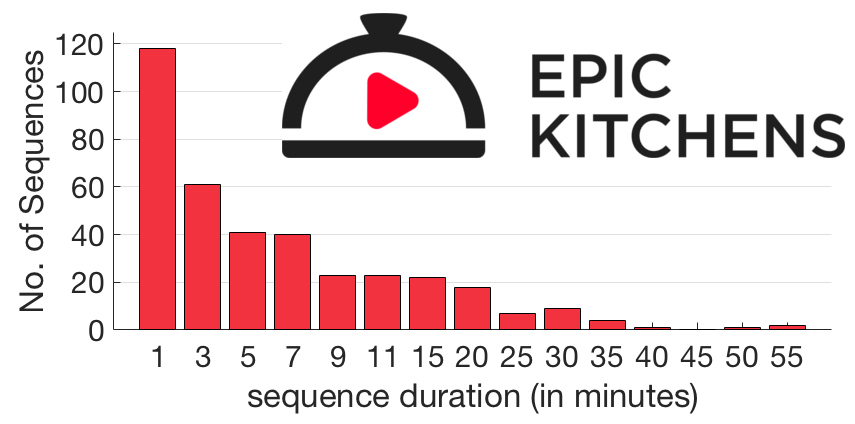}\\
{\includegraphics[width=0.2\columnwidth]{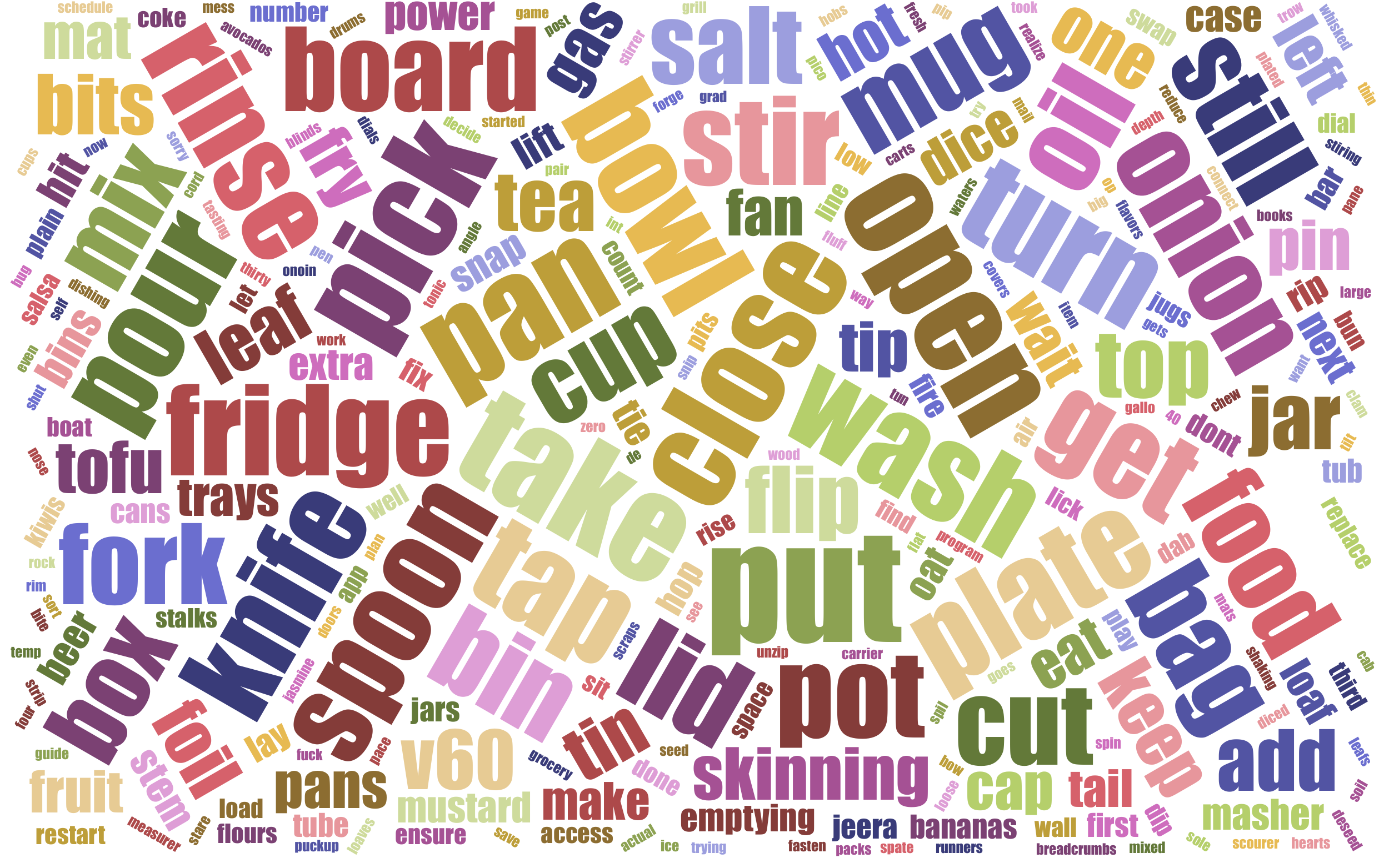}}
\subfloat 
{\includegraphics[width=0.2\columnwidth]{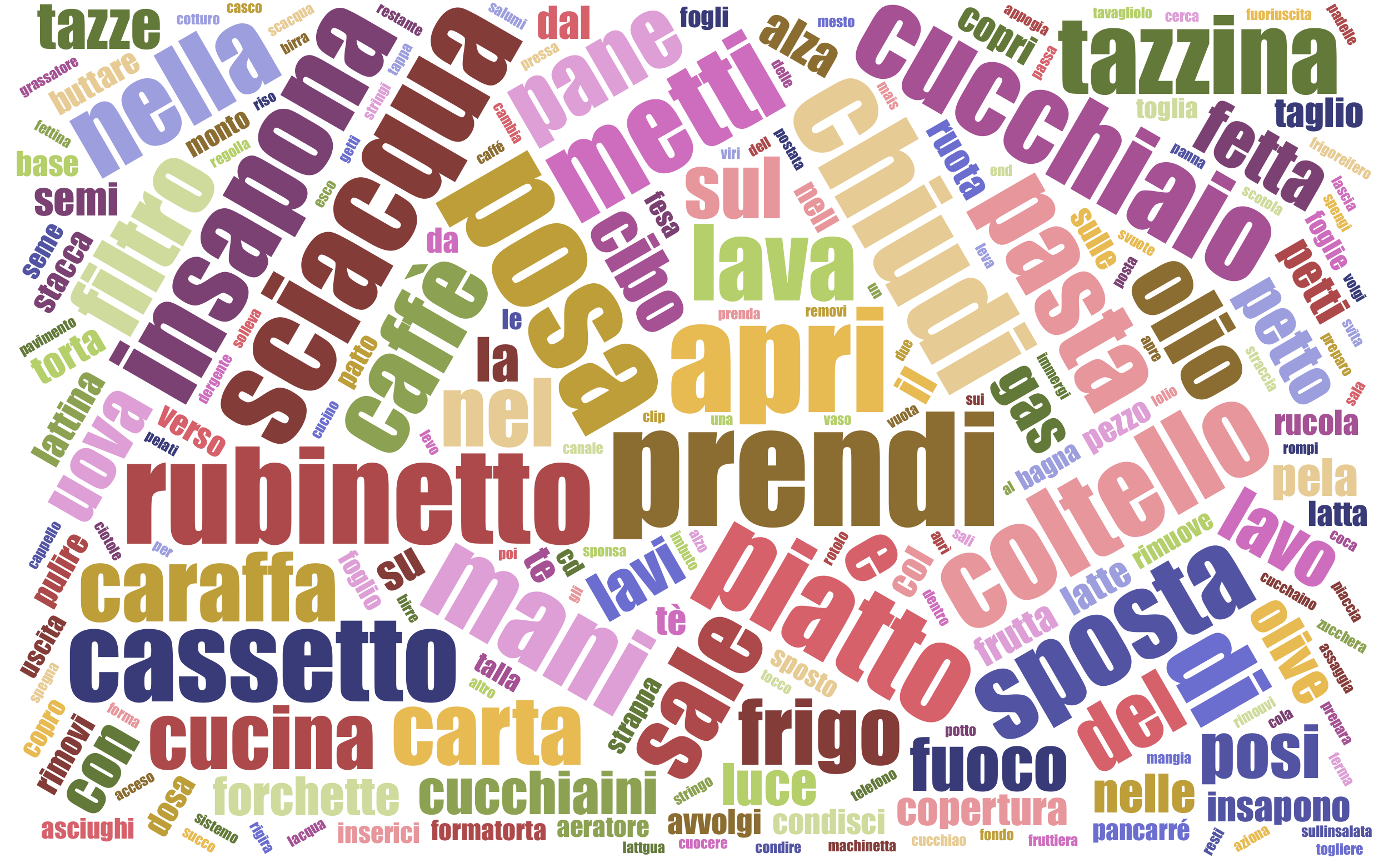}}
\subfloat 
{\includegraphics[width=0.2\columnwidth]{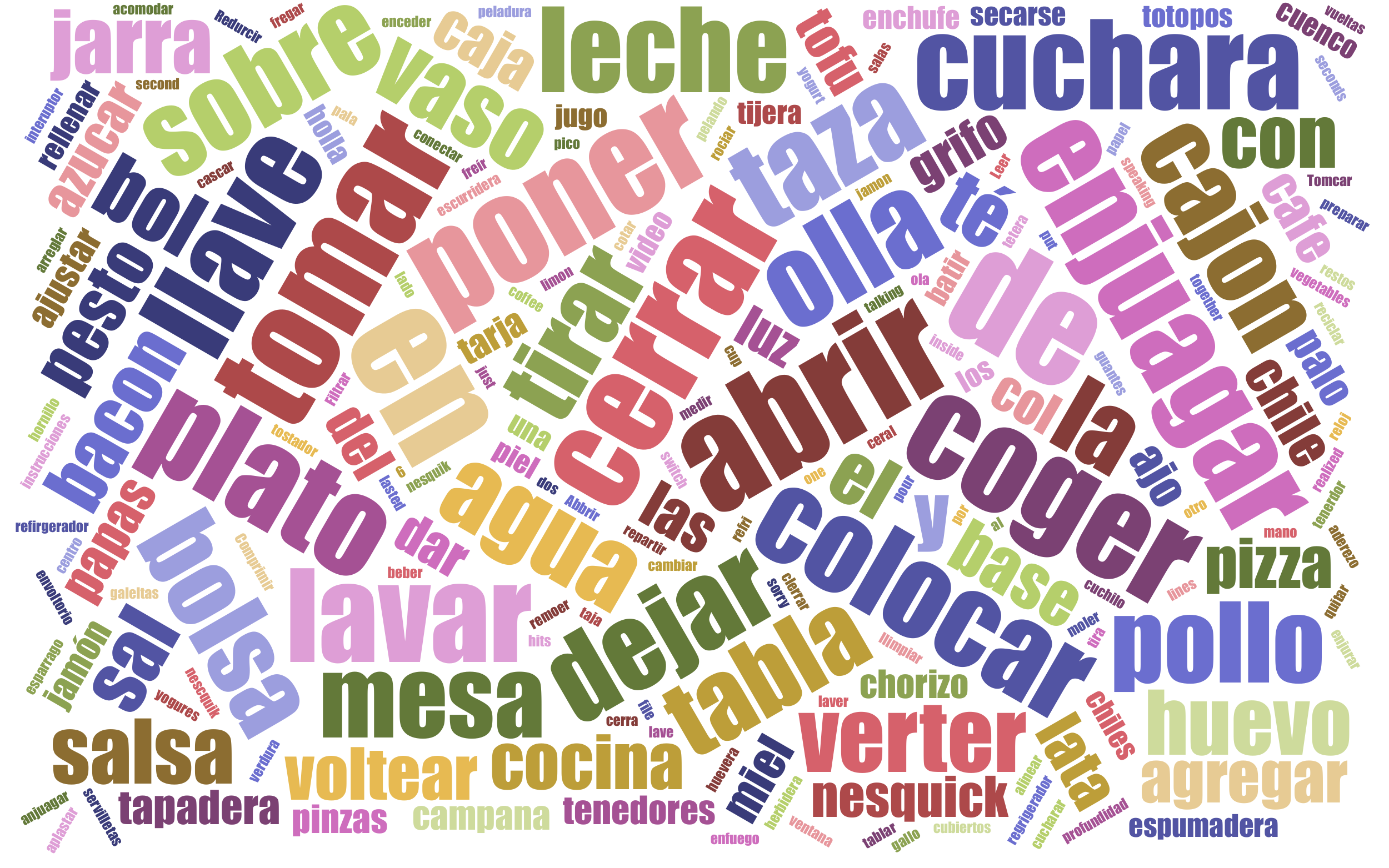}}
\subfloat
{\includegraphics[width=0.2\columnwidth]{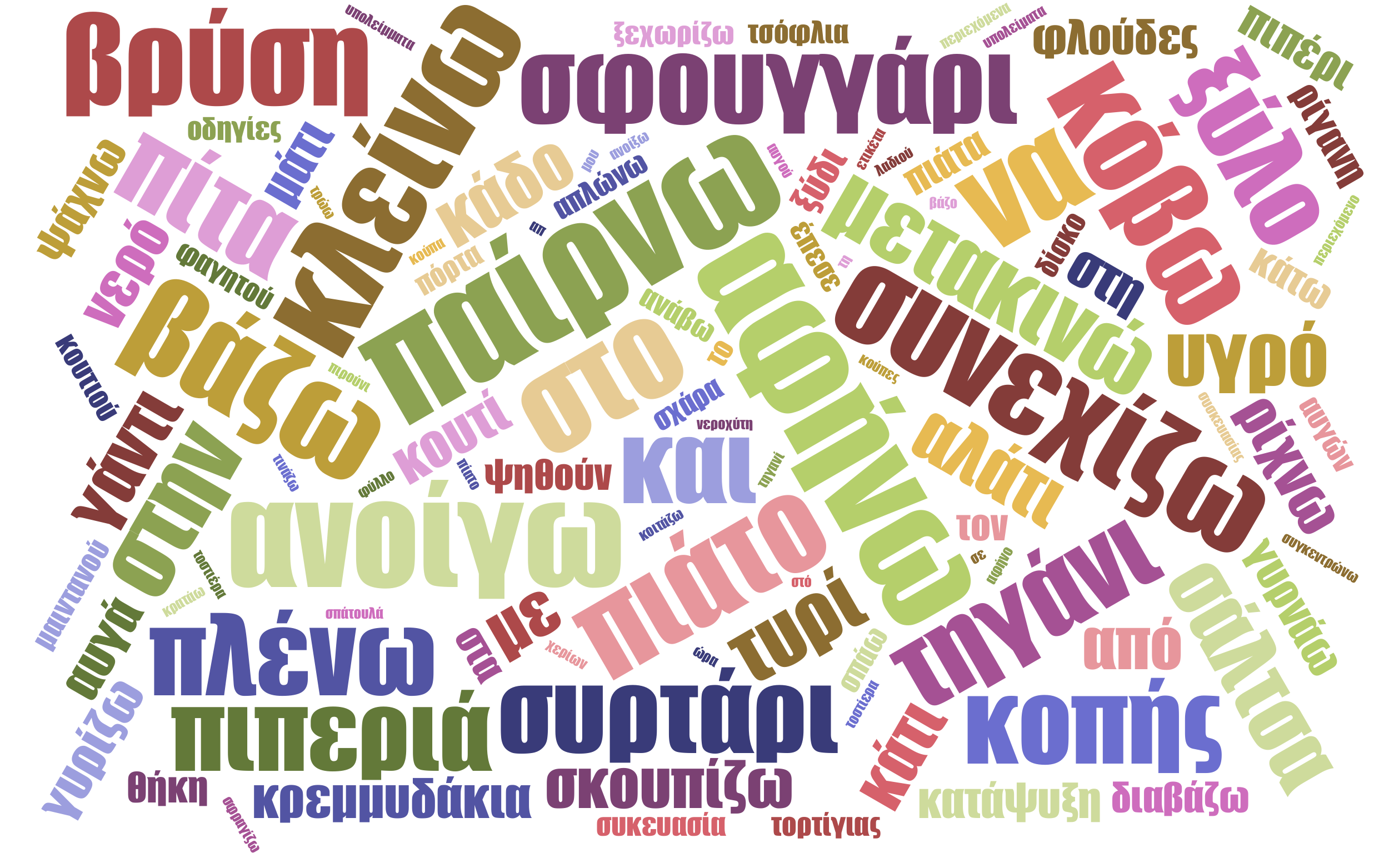}}
\subfloat
{\includegraphics[width=0.2\columnwidth]{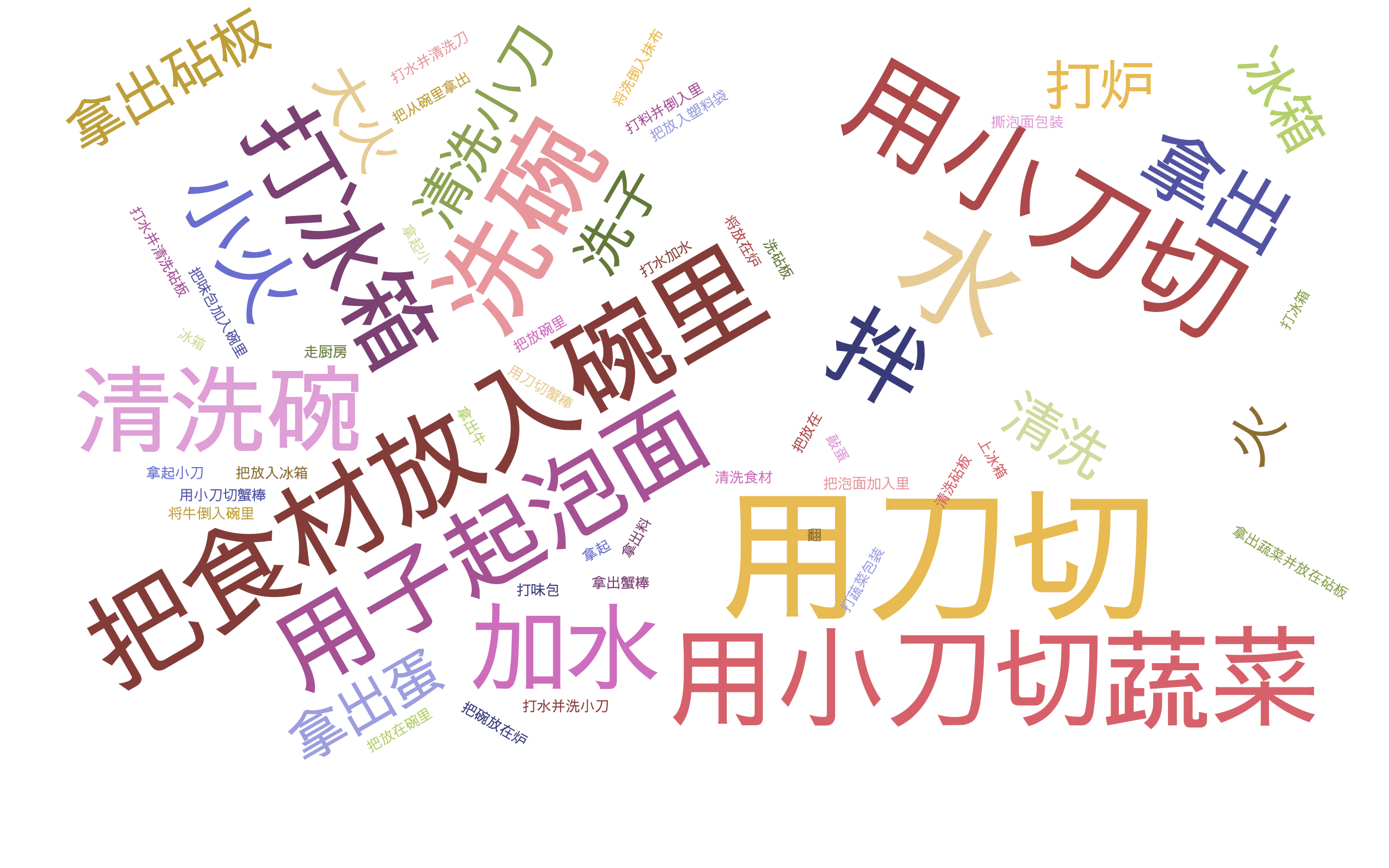}}
\vspace{-2mm}
\caption{{\bf Top} (left to right): time of day of the recording, pie chart of high-level goals, histogram of sequence durations and dataset logo; {\bf Bottom}: Wordles of narrations in native languages (English, Italian, Spanish, Greek and Chinese)}
\label{fig:statsCollect}
\vspace{-3mm}
\end{figure}

The recording lengths varied depending on the participant's kitchen engagement.  
On average, people recorded for 1.7hrs, with the maximum being 4.6hrs. Cooking a single meal can span multiple sequences, depending on whether one stays in the kitchen, or leaves and returns later. On average, each participant recorded 13.6 sequences.
Figure~\ref{fig:statsCollect} presents statistics on time of day using the local time of the recording, high-level goals and sequence durations.

Since crowd-sourcing annotations for such long videos is very challenging, we had our original participants do a coarse first annotation. 
Each participant was asked to watch their videos, after completing all recordings, and narrate the actions carried out, using a hand-held recording device. We opted for a sound recording rather than written captions as this is arguably much faster for the participants, who were thus more willing to provide these annotations. These are analogous to a \textit{live commentary} of the video. The general instructions for narrations are listed in Fig.~\ref{fig:narrationInstruction}. The participant narrated in English if sufficiently fluent or in their native language. In total, 5 languages were used: 17 narrated in English, 7 in Italian, 6 in Spanish, 1 in Greek and 1 in Chinese. Figure~\ref{fig:statsCollect} shows wordles of the most frequent words in each language.

\begin{table}[t]
\caption{Extracts from 6 transcription files in .sbv format}
\resizebox{\textwidth}{!}{%
\begin{tabular}{ |l|l|l|l|l|l| } 
\hline
 0:14:44.190,0:14:45.310 & 0:00:02.780,0:00:04.640 & 0:04:37.880,0:04:39.620 & 0:06:40.669,0:06:41.669 & 0:12:28.000,0:12:28.000 & 0:00:03.280,0:00:06.000 \\ 
 pour tofu onto pan & open the bin & Take onion & pick up spatula & pour pasta into container & open fridge\\ 
 0:14:45.310,0:14:49.540 & 0:00:04.640,0:00:06.100 & 0:04:39.620,0:04:48.160 & 0:06:41.669,0:06:45.250 & 0:12:33.000,0:12:33.000 & 0:00:06.000,0:00:09.349 \\
 put down tofu container & pick up the bag & Cut onion & stir potatoes & take jar of pesto & take milk \\
 0:14:49.540,0:15:02.690 & 0:00:06.100,0:00:09.530 & 0:04:48.160,0:04:49.160 & 0:06:45.250,0:06:46.250 &0 :12:39.000,0:12:39.000 & 0:00:09.349,0:00:10.910 \\
 stir vegetables and tofu & tie the bag & Peel onion & put down spatula & take teaspoon & put milk \\
 0:15:02.690,0:15:06.260 & 0:00:09.530,0:00:10.610 & 0:04:49.160,0:04:51.290 & 0:06:46.250,0:06:50.830& 0:12:41.000,0:12:41.000 &0:00:10.910,0:00:12.690  \\
 put down spatula & tie the bag again & Put peel in bin & turn down hob & pour pesto in container & open cupboard \\
 0:15:06.260,0:15:07.820 & 0:00:10.610,0:00:14.309 & 0:04:51.290,0:05:06.350 & 0:06:50.830,0:06:55.819 & 0:12:55.000,0:12:55.000 & 0:00:12.690,0:00:15.089 \\
 take tofu container & pick up bag &  Peel onion & pick up pan & place pesto bottle on table &take bowl \\
 0:15:07.820,0:15:10.040 & 0:00:14.309,0:00:17.520 & 0:05:06.350,0:05:15.200 & 0:06:55.819,0:06:57.170 & 0:12:58.000,0:12:58.000 & 0:00:15.089,0:00:18.080 \\
 throw something into the bin & put bag down & Put peel in bin & tip out paneer & take wooden spoon & open drawer \\
 \hline
\end{tabular}}
\label{fig:transcriptfiles}
\vspace{-2mm}
\end{table}

Our decision to collect narrations from the participants themselves is because they are the most qualified to label the activity compared to an independent observer, as they were the ones performing the actions. We opted for a post-recording narration such that the participant performs her/his daily activities undisturbed, without being concerned about labelling.

We tested several automatic audio-to-text APIs \cite{googleSpeechApi,ibmWatson,cmuSphinx}, which failed to produce accurate transcriptions as these expect a relevant corpus and complete sentences for context. We thus collected manual transcriptions via Amazon Mechanical Turk (AMT), and used the YouTube's automatic closed caption alignment tool to produce accurate timings. For non-English narrations, we also asked AMT workers to translate the sentences. 
To make the job more suitable for AMT, narration audio files are split by removing silence below a pre-specified decibel threshold (after compression and normalisation).  Speech chunks are then combined into HITs with a duration of around 30 seconds each.
To ensure consistency, we submit the same HIT three times and select the ones with an edit distance of 0 to at least one other HIT. We manually corrected cases when there was no agreement. Examples of transcribed and timed narrations are provided in Table~\ref{fig:transcriptfiles}.
The participants were also asked to provide one sentence per sequence describing the overall goal or activity that took place. 

In total, we collected $39,596$ action narrations, corresponding to a narration every $4.9s$ in the video. The average number of words per phrase is $2.8$ words.
These narrations give us an initial labelling of all actions with rough temporal alignment, obtained from the timestamp of the audio narration with respect to the video. However, narrations are also not a perfect source of ground-truth:

\begin{itemize}[leftmargin=*]
\item The narrations can be incomplete, i.e., the participants were selective in which actions they chose to narrate. We noticed that they labelled the `open' actions more than their counter-action `close', as the narrator's attention has already moved to the next goal. We consider this phenomena in our evaluation, by only evaluating actions that have been narrated.

\item Temporally, 
the narrations are belated, after the action takes place. This is adjusted using ground-truth action segments (see Sec.~\ref{subsec:strongAction}).
\item Participants use their own vocabulary and free language. While this is a challenging issue, we believe it is important to push the community to go beyond the pre-selected list of labels (also argued in~\cite{ade}). We here resolve this issue by grouping verbs and nouns into minimally overlapping classes (see Sec.~\ref{subsec:clusters}).
\end{itemize}

\subsection{Action Segment Annotations}
\label{subsec:strongAction}

For each narrated sentence, we adjust the start and end times of the action using AMT. 
To ensure the annotators are trained to perform temporal localisation, we use a clip from our previous work's understanding~\cite{Moltisanti2017} that explains temporal bounds of actions.
Each HIT is composed of a maximum of 10 consecutive narrated phrases $p_i$, where annotators label $A_{i} = [t_{s_i}, t_{e_i}]$ as the start and end times of the $i^{th}$ action. Two constraints were added to decrease the amount of noisy annotations: (1) action has to be at least 0.5 seconds in length; (2) action cannot start before the preceding action's start time. Note that consecutive actions are allowed to overlap.
Moreover, the annotators could indicate that the action does not appear in the video. This handles occluded, impossible to distinguish or out-of-bounds cases.

\begin{figure}[t]
\centering
\includegraphics[width=\linewidth]{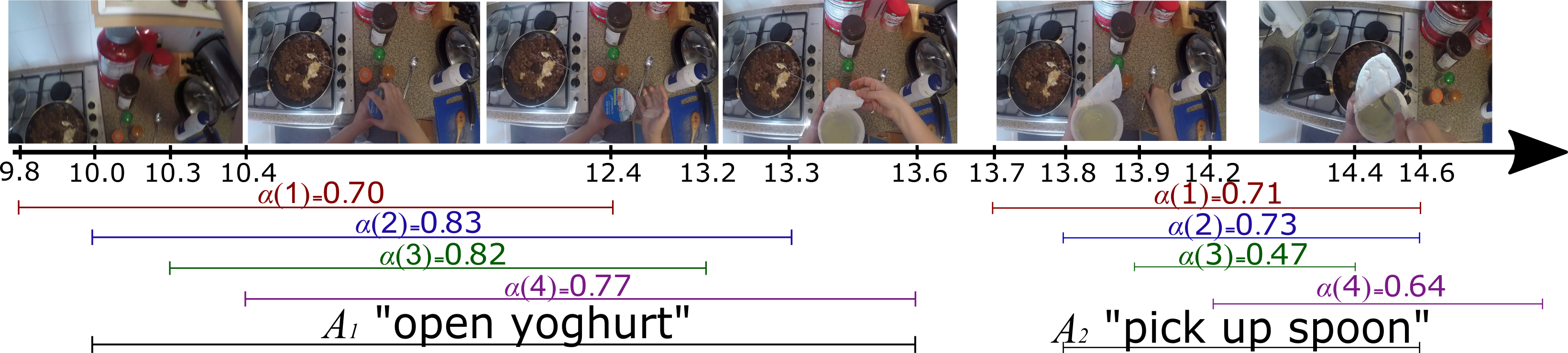}
\caption{An example of annotated action segments for 2 consecutive actions}
\label{fig:annotations_example}
\vspace*{10pt}

\includegraphics[width=\linewidth]{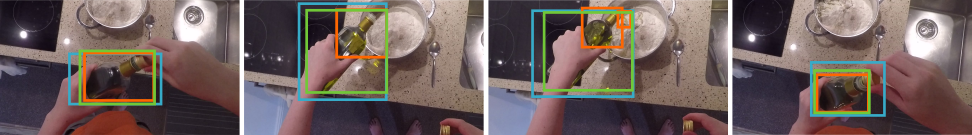}
\caption{Object annotation from three AMT workers (orange, blue and green). The green participant's annotations are selected as the final annotations}
\label{fig:object_ann}
\end{figure}

To ensure consistency, we ask $\mathcal{K}_a=4$ annotators to annotate each HIT. Given one annotation $A_i(j)$ ($i$ is the action and $j$ indexes the annotator), we calculate the agreement as follows: ${\alpha_i(j) = \frac{1}{K_a} \sum_{k=1}^{\mathcal{K}_a} \text{IoU} (A_i(j), A_i(k))}$. We first find the annotator with the maximum agreement $\hat{j} = \arg\max_j \alpha_i(j)$, and find $\hat{k} = \arg\max_k \text{IoU}(A_i(\hat{j}), A_i(k))$. The ground-truth action segment $A_i$ is then defined as:
\begin{equation}
\small{
A_i =\begin{cases}
\text{Union}(A_i(\hat{j}), A_i(\hat{k})), & \text{if  IoU} (A_i(\hat{j}), A_i(\hat{k}))>0.5\\
A_i(\hat{j}), & \text{otherwise}
\end{cases}}
\label{eq:actionunion}
\end{equation}
We thus combine two annotations when they
have a strong agreement, since in some cases the single (best) annotation results in a too tight of a segment. Figure~\ref{fig:annotations_example} shows examples of combining annotations.

In total, we collected such labels for $39,564$ action segments (lengths: ${\mu=3.7s}$, ${\sigma=5.6s}$). These represent 99.9\%  of narrated segments. The missed annotations were those labelled as ``not visible'' by the annotators, though mentioned in narrations. 

\subsection{Active Object Bounding Box Annotations}
\label{subsec:strongObject}

The narrated \textit{nouns} correspond to objects relevant
to the action~\cite{Lee2012,Damen2014a}. Assume $\mathcal{O}_i$ is the set of one or more nouns in the phrase $p_i$ associated with the action segment $A_i = [t_{s_i}, t_{e_i}]$. We consider each frame $f$ within $[t_{s_i}-2s,t_{e_i}+2s]$ as a potential frame to annotate the bounding box(es), for each object in $\mathcal{O}_i$. We build on the interface from~\cite{tangseng2017} for annotating bounding boxes on AMT.
Each HIT aims to get an annotation for one object, for the maximum duration of $25s$, which corresponds to $50$ consecutive frames at $2$\fps. The annotator can also note that the object does not exist in~$f$. 
We particularly ask the same annotator to annotate consecutive frames to avoid subjective decisions on the extents of objects. We also assess annotators' quality by ensuring that the annotators obtain an ${\textrm{IoU} \ge 0.7}$ on two golden annotations at the start of every HIT. We request $\mathcal{K}_o = 3$ workers per HIT, and select the one with maximum agreement $\beta$: 
\begin{equation}
\beta(q) = \sum_f \max\limits_{j \ne q}^{\mathcal{K}_o}\, \max_{k,l}\, \text{IoU}(\text{BB}(j, f, k), \text{BB}(q, f, l))
\end{equation}

\vspace*{-8pt}
\noindent where $\text{BB}(q,f, k)$ is the $k^{th}$ bounding box annotation by annotator $q$ in frame~$f$. Ties are broken by selecting the worker who provides the tighter bounding boxes. Figure~\ref{fig:object_ann} shows multiple annotations for four keyframes in a sequence. 

\begin{figure}
\centering
\includegraphics[width=\linewidth]{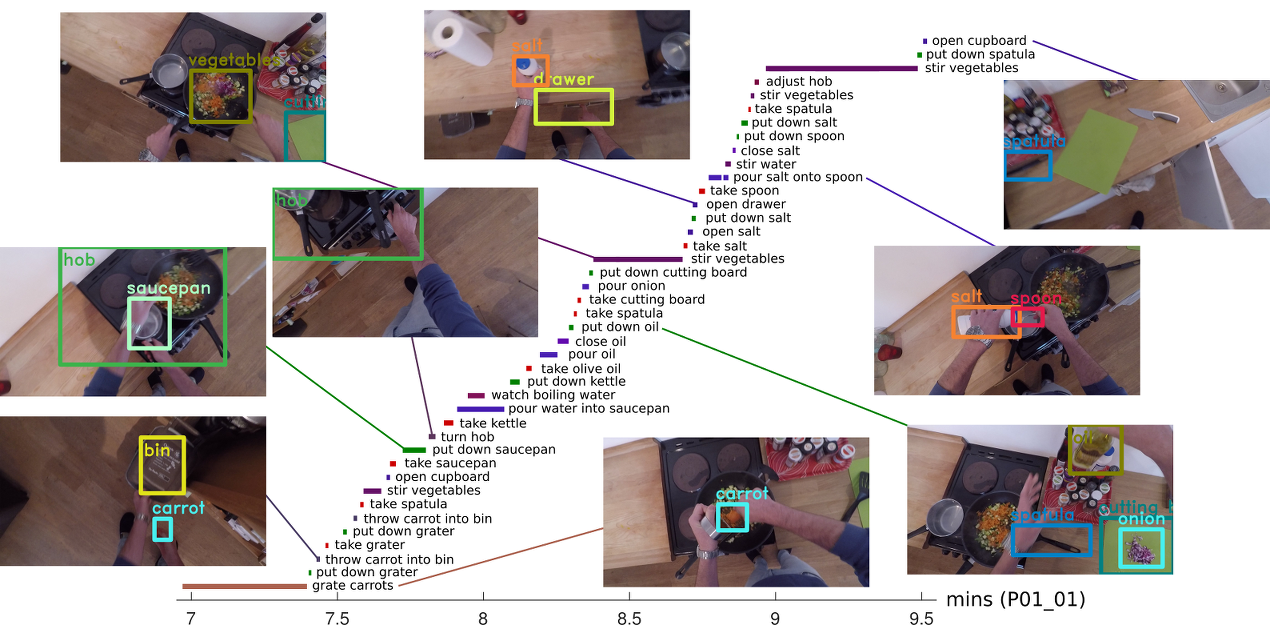}
\caption{Sample consecutive action segments with keyframe object annotations}
\label{fig:timeline_example}
\end{figure}

Overall, 77\% of requested annotations resulted in at least one bounding box.
In total, we collected 454,255 bounding boxes ($\mu = 1.64$ boxes/frame, ${\sigma = 0.92}$). Sample action segments and object bounding boxes are shown in Fig.~\ref{fig:timeline_example}.

\subsection{Verb and Noun Classes}
\label{subsec:clusters}

Since our participants annotated using free text in multiple languages, a variety of verbs and nouns have been collected. We group these into classes with minimal semantic overlap, to accommodate the more typical approaches to multi-class detection and recognition where each example is believed to belong to one class only. We estimate Part-of-Speech (POS), using SpaCy's English core web model. We select the first verb in the sentence, and find all nouns in the sentence excluding any that match the chosen verb.
When a noun is absent or replaced by a pronoun (\textit{e.g. `it'}), we use the noun from the directly preceding narration (e.g. $p_i$: `rinse cup', $p_{i+1}$: `place it to dry').

\begin{figure*}[t]
\includegraphics[width=1\textwidth]{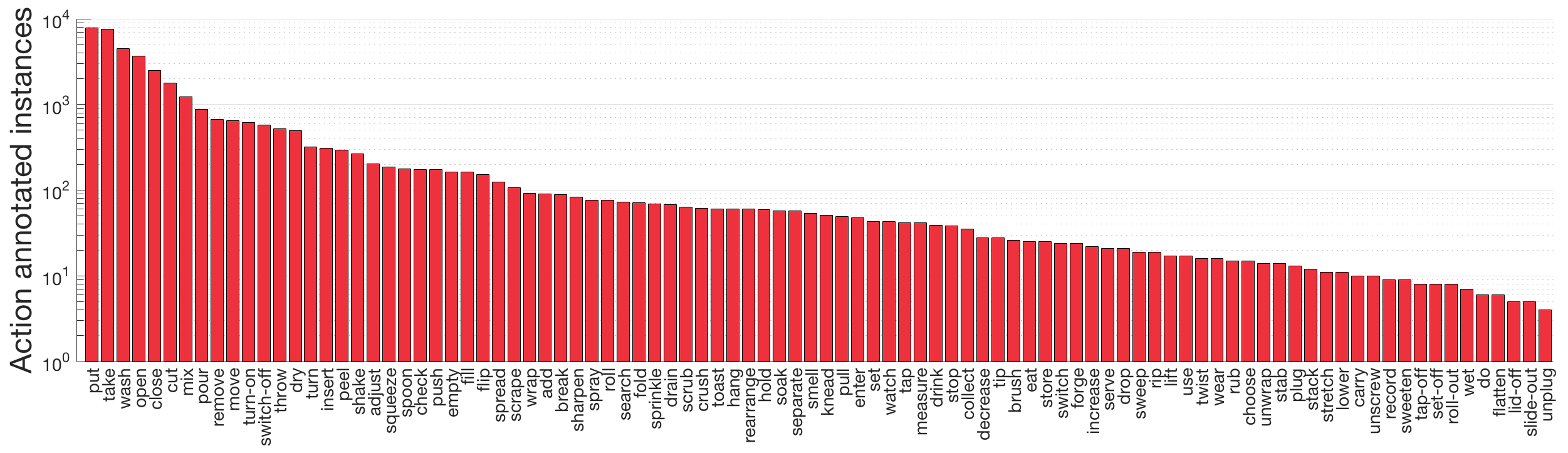}\\
\includegraphics[width=1\textwidth]{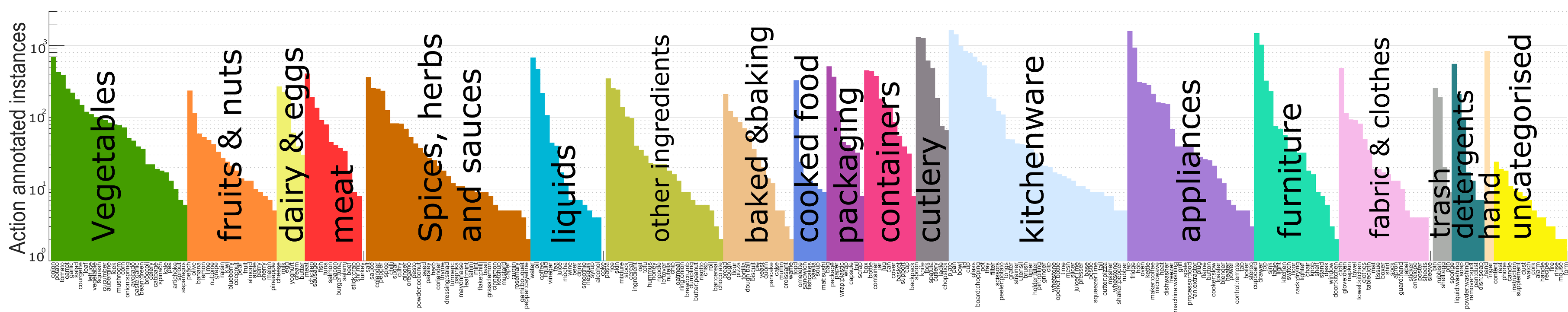}\\
\includegraphics[width=1\textwidth]{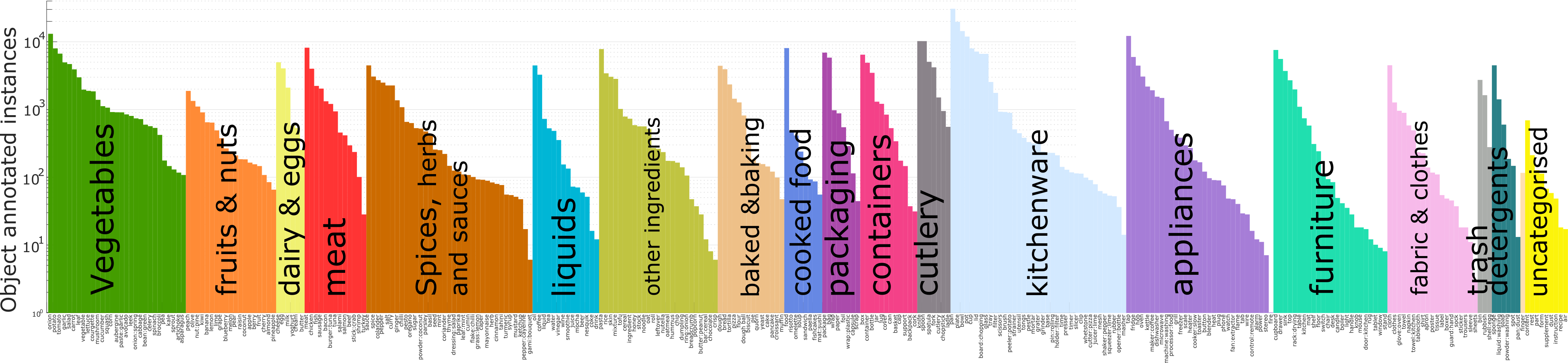}\\
\includegraphics[width=1\textwidth]{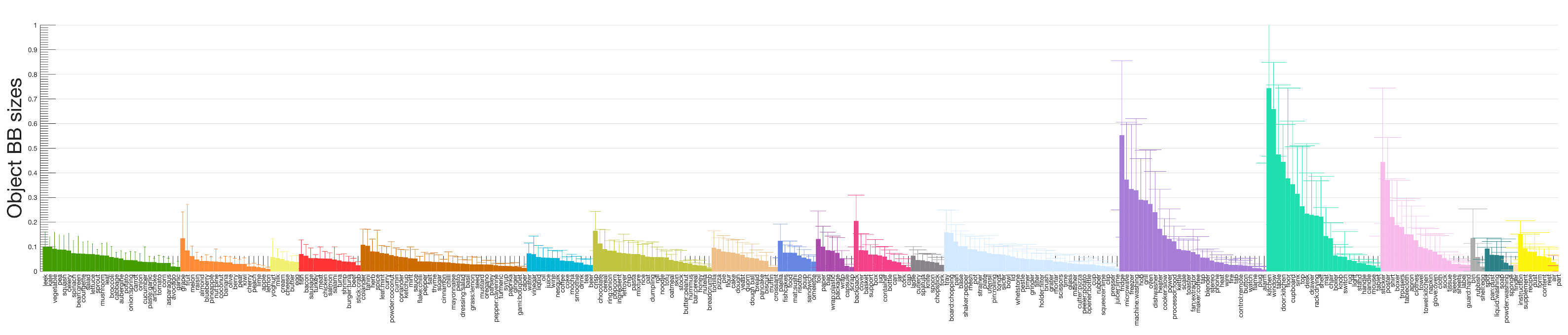}
\caption{{\bf From Top}: Frequency of verb classes in action segments; Frequency of noun clusters in action segments, by category; Frequency of noun clusters in bounding box annotations, by category; Mean and standard deviation of bounding box, by category}
\label{fig:stats}
\end{figure*}

We refer to the set of minimally-overlapping verb classes as $C_V$, and similarly $C_N$ for nouns.
We attempted to automate the clustering of verbs and nouns using combinations of WordNet~\cite{miller1995wordnet}, Word2Vec~\cite{mikolov2013efficient}, and Lesk algorithm~\cite{Banerjee2002}, however, due to limited context there were too many meaningless clusters. We thus elected to manually cluster the verbs and semi-automatically cluster the nouns. We preprocessed the compound nouns \textit{e.g.}~`pizza cutter' as a subset of the second noun \textit{e.g.}~`cutter'. We then manually adjusted the clustering, merging the variety of names used for the same object, \textit{e.g.} `cup' and `mug', as well as splitting some base nouns, \textit{e.g.}~`washing machine' vs `coffee machine'.

\begin{figure}[t]
\vspace{-1mm}
{\includegraphics[width=0.52\columnwidth]{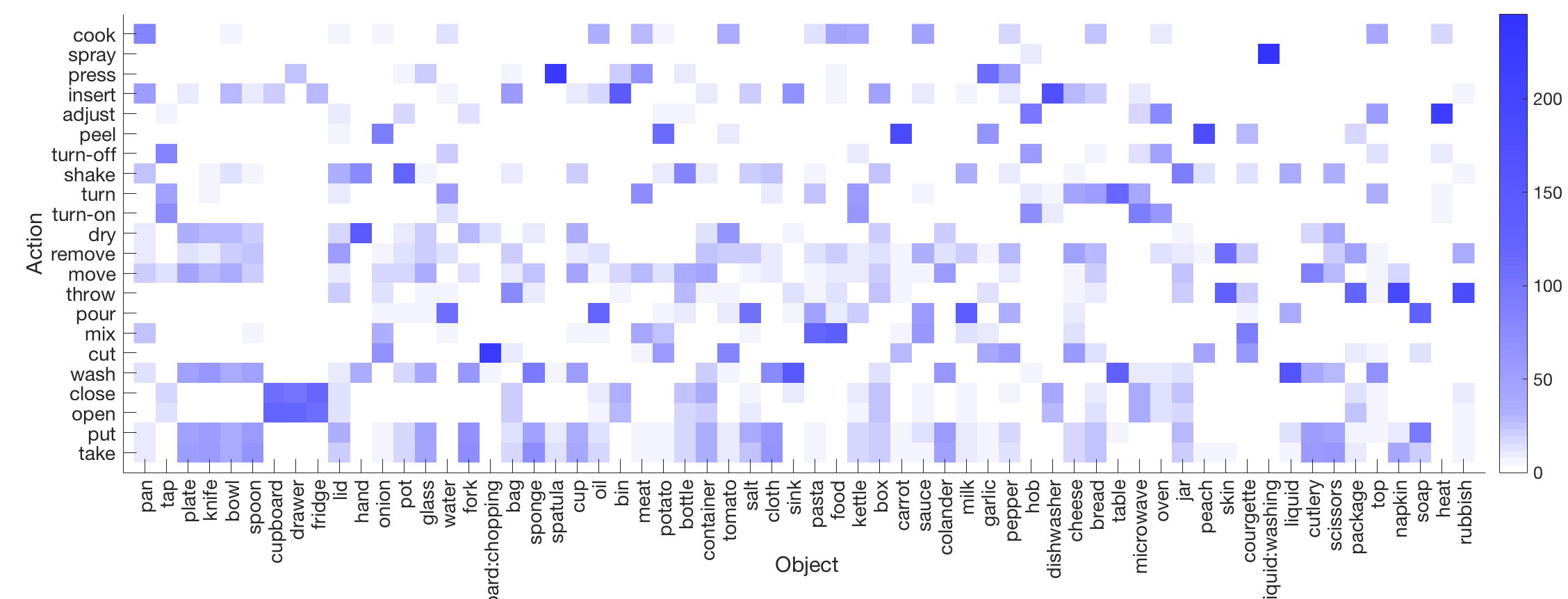}}
{\includegraphics[width=0.23\columnwidth]{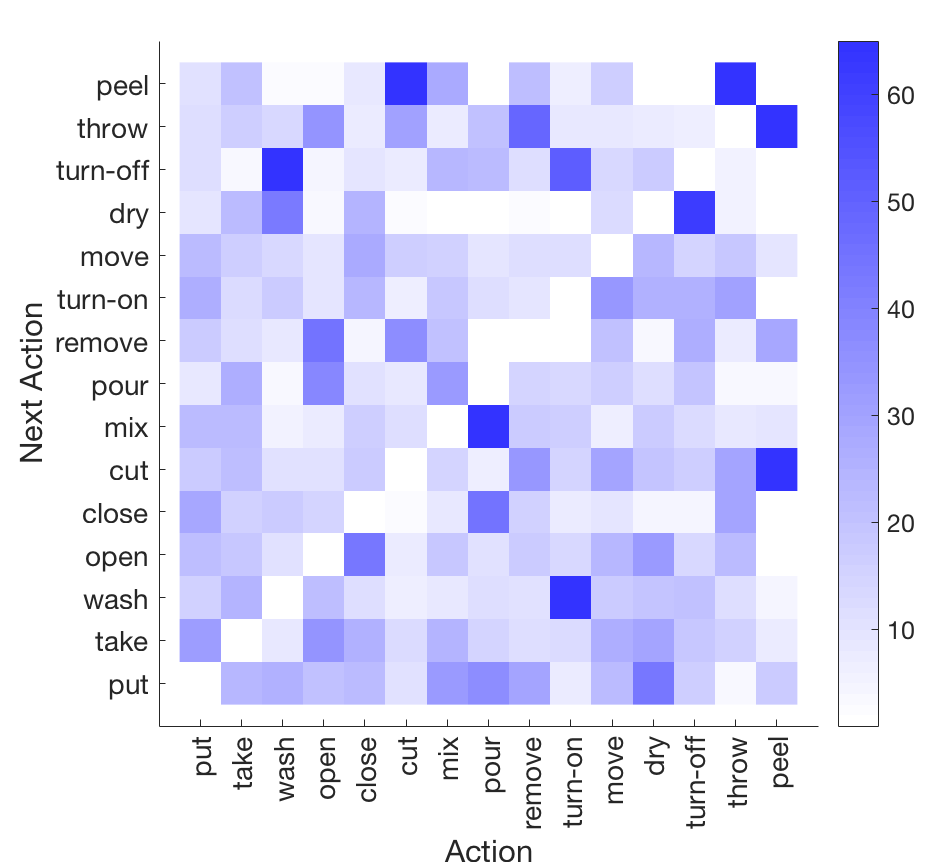}}
{\includegraphics[width=0.23\columnwidth]{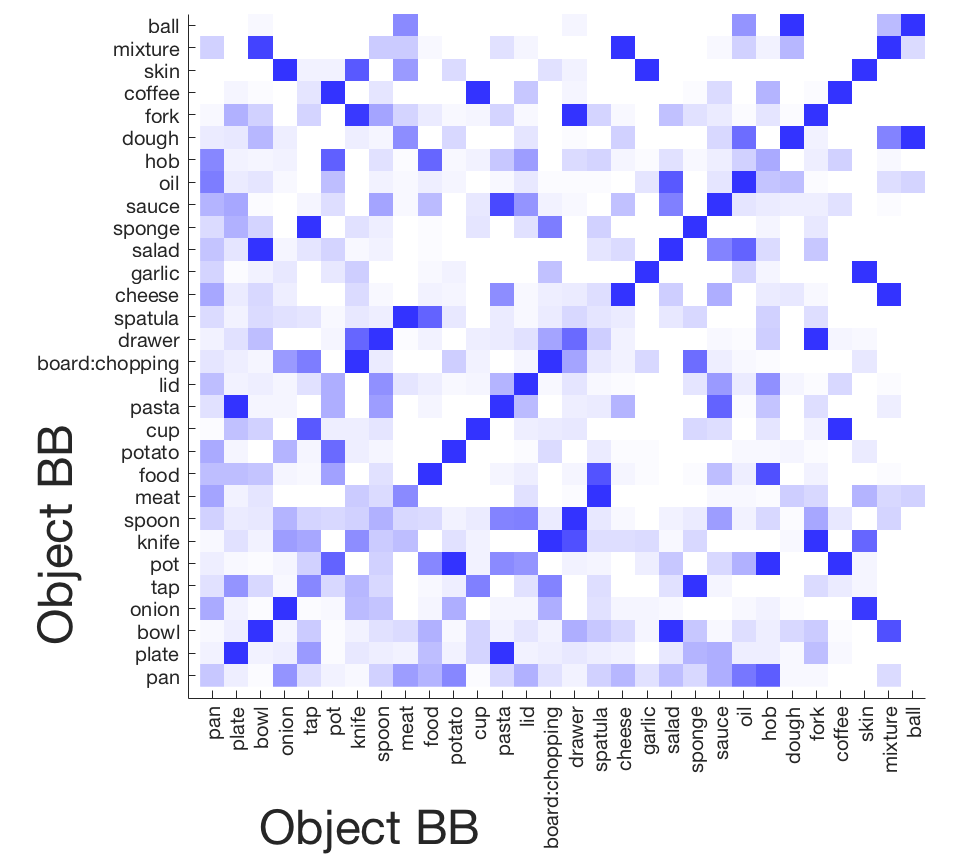}}
\vspace{-5mm}
\caption{\small {\bf Left}: Frequently co-occurring verb/nouns in action segments [e.g. (open/close, cupboard/drawer/fridge), (peel, carrot/onion/potato/peach), (adjust, heat)]; {\bf Middle}:~Next-action excluding repetitive instances of the same action [e.g. peel $\rightarrow$ cut, turn-on $\rightarrow$ wash, pour $\rightarrow$ mix].; {\bf Right}: Co-occurring bounding boxes in one frame [e.g.~(pot, coffee), (knife, chopping board), (tap, sponge)]}
\label{fig:co-occur}
\end{figure}

In total, we have 125 $C_V$ classes and 331 $C_N$ classes. Table~\ref{tab:classes} shows a sample of grouped verbs and nouns into classes.
These classes are used in all three defined challenges.
In Fig.~\ref{fig:stats}, we show $C_V$ ordered by frequency of occurrence in action segments, as well as $C_N$ ordered by number of annotated bounding boxes. These are grouped into 19 super categories, of which 9 are food and drinks, with the rest containing kitchen essentials from appliances to cutlery. Co-occurring classes are presented in Fig.~\ref{fig:co-occur}.

\subsection{Annotation Quality Assurance}
To analyse the quality of annotations, we choose  300 random samples, and manually assess correctness. We report:

\begin{itemize}[leftmargin=*]
\item \textbf{Action Segment Boundaries ($A_i$)}: We check that the start/end times fully enclose the action boundaries, with any additional frames not part of other actions - error: \textit{5.7\%}.
\item \textbf{Object Bounding Boxes ($\mathcal{O}_i$)}: We check that the bounding box encapsulates the object or its parts, with minimal overlap with other objects, and that all instances of the class in the frame have been labelled -- error: \textit{6.3\%}. 
\item \textbf{Verb classes ($C_V$):} We check that the verb class is correct -- error: \textit{3.3\%}.
\item \textbf{Noun classes ($C_N$):} We check that the noun class is correct -- error : \textit{6.0\%}.

\end{itemize}
\vspace*{-3pt}

\noindent These error rates are comparable to recently published datasets~\cite{zhao2017slac}.

\section{Benchmarks and Baseline Results}
\label{sec:baselines}

\EPIC{} offers a variety of potential challenges from routine understanding, to activity recognition and object detection. As a start, 
we define three challenges for which we provide baseline results, and avail online leaderboards.
For the evaluation protocols, we hold out ground truth annotations for 27\% of the data (Table~\ref{tab:s1_stats}).
We particularly aim to assess the generalizability to novel environments, and we thus structured our test set to have a collection of \textit{seen} and previously \textit{unseen} kitchens:

\noindent\textbf{Seen Kitchens (S1):} In this split,
each kitchen is seen in both training and testing, where roughly 80\% of sequences are in training and 20\% in testing. We do not split sequences, thus each sequence is in either training or testing. 

\noindent\textbf{Unseen Kitchens (S2):} This divides the participants/kitchens so all sequences of the same kitchen are either in training or testing. We hold out the complete sequences for 4 participants for this testing protocol. The test set of S2 is only 7\% of the dataset in terms of frame count, but the challenges remain considerable. 

\begin{table}[t!]
\vspace{-1mm}
\centering
\caption{Sample Verb and Noun Classes}
\resizebox{0.9\textwidth}{!}{%
\begin{tabular}{l|l|l} 
&ClassNo (Key) &Clustered Words\\
\hline
\multirow{3}{*}{\rotatebox{90}{\textbf{VERB}}} &0 (take) &take, grab, pick, get, fetch, pick-up, ...\\
&3 (close) &close, close-off, shut\\
&12	(turn-on) &turn-on, start, begin, ignite, switch-on, activate, restart, light, ...\\ \hline
\multirow{4}{*}{\rotatebox{90}{\textbf{NOUN}}} &1 (pan) &pan, frying pan, saucepan, wok, ...\\
&8 (cupboard) & cupboard, cabinet, locker, flap, cabinet door, cupboard door, closet, ...\\
&51 (cheese) &cheese slice, mozzarella, paneer, parmesan, ...\\
&78 (top) &top, counter, counter top, surface, kitchen counter, kitchen top, tiles, ...
\end{tabular}}
\label{tab:classes}

\caption{Statistics of test splits: seen (S1) and unseen (S2) kitchens}
\label{tab:s1_stats}
\resizebox{\columnwidth}{!}{%
\begin{tabular}{|l|c|c|c|c|c|c|c|}
\hline
& \#Subjects & \#Sequences & Duration (s) & \% &Narrated Segments &Action Segments &Bounding Boxes\\
\hline\hline
Train/Val &28 &272 &141731 & &28,587 &28,561 &326,388
\\ \hline
 \textbf{S1} Test &28 &106 &39084 &20\% &8,069 &8,064 &97,872\\
\hline
\textbf{S2} Test &4 &54 &13231 &7\% &2,939 &2,939 &29,995\\
\hline
\end{tabular}}
\vspace{-4mm}
\end{table}

We now evaluate several existing methods on our benchmarks, to gain an understanding of how challenging our dataset is. 

\vspace{-6mm}
\subsection{Object Detection Benchmark}
\label{subsec:obj_challenges}
\vspace*{-2mm}
\noindent\textbf{Challenge:}
This challenge focuses on object detection for all of our $C_N$ classes. 
Note that our annotations only capture the `active' objects pre-, during- and post- interaction. 
We thus restrict the images evaluated per class to those where the object has been annotated.
We particularly aim to break the performance down into multi-shot and few-shot class groups, so as to analyse the capabilities of the approaches to quickly learn novel objects (with only a few examples). Our challenge leaderboard reflects the methods' abilities on both sets of classes. 

\noindent\textbf{Method:} We evaluate object detection using Faster R-CNN~\cite{ren2015faster} due to its state-of-the-art performance. Faster R-CNN uses a region proposal network (RPN) to first generate class agnostic object proposals, and then classifies these and outputs refined bounding box predictions. We use the implementation from~\cite{objapi,huang2017speed} with a base architecture of ResNet-101~\cite{resnet} pre-trained on MS-COCO~\cite{coco}.

\noindent\textbf{Implementation Details:} Learning rate is initialised to 0.0003 decaying by a factor of 10 after 90K and stopped after 120K iterations.  We use a mini-batch size of 4 on 8 Nvidia P100 GPUs on a single compute node (Nvidia DGX-1) with distributed training and parameter synchronisation -- i.e. overall mini-batch size of 32. As in~\cite{ren2015faster}, images are rescaled such that their shortest side is 600 pixels and the aspect ratio is maintained. We use a stride of 16 on the last convolution layer for feature extraction and for anchors we use 4 scales of 0.25, 0.5, 1.0 and 2.0; and aspect ratios of 1:1, 1:2 and 2:1. To reduce redundancy, NMS is used with an IoU threshold of 0.7. In training and testing we use 300 RPN proposals.

\noindent\textbf{Evaluation Metrics:}
For each class, we only report results on $I^{c_n \in C_N}$, these are all images where class $c_n$ has been annotated. We use the mean average precision (mAP) metric from PASCAL VOC~\cite{pascal}, using IoU thresholds of 0.05, 0.5 and 0.75 similar to~\cite{coco}. 

\noindent\textbf{Results:} We report results in Table~\ref{tab:results_obj} for many-shot classes (those with ${\geq 100}$ bounding boxes in training) and few shot classes (with ${\geq 10}$ and ${< 100}$ bounding boxes in training), alongside AP for the 15 most frequent classes. There are a total of 202 many-shot classes and 88 few-shot classes. One can see that our objects are generally harder to detect than in most existing datasets, with performance at the standard IoU $>0.5$ below $40\%$. Even at a very small IoU threshold, the performance is relatively low. The more challenging classes are ``meat'', ``knife'', and ``spoon'', despite being some of the most frequent ones. Notice that the performance for the low-shot regime is substantially lower than in the many-shot regime. This points to interesting challenges for the future. However, performances for the \emph{Seen} and \emph{Unseen} splits in object detection are comparable, thus showing generalization capability across environments. 

\begin{figure}[t!]
\vspace{-2mm}
\includegraphics[width=1\linewidth,height=5cm]{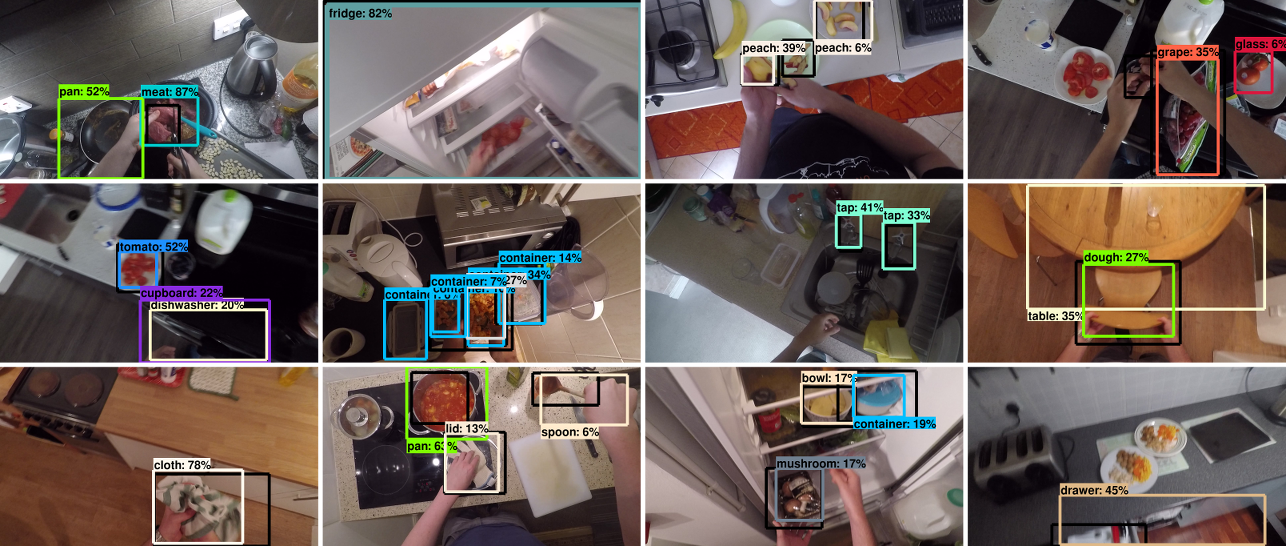}
\vspace{-6mm}
\caption{Qualitative results for the object detection challenge}
\label{fig:qual_obj}
\end{figure}

Figure~\ref{fig:qual_obj} shows qualitative results  with detections shown in colour and ground truth shown in black. The examples in the right-hand column are failure cases.

\begin{table}[t!]
\vspace{-2.0mm}
\centering
\caption{Baseline results for the Object Detection challenge}
\resizebox{\columnwidth}{!}{%
\begin{tabular}{@{}ll|ccccccccccccccc|c|c|c@{}}
& &\multicolumn{15}{c|}{\textbf{15 Most Frequent Object Classes}} &\multicolumn{3}{c}{\textbf{Totals}}\\ \cline{3-20}
 &mAP & pan & plate & bowl & onion & tap & pot & knife & spoon & meat & food & potato & cup & pasta & cupboard & lid & few-shot & many-shot & all\\\hline
 \multirow{3}{*}{\rotatebox{90}{\textbf{S1}}} & IoU $>0.05$ & 78.40 & 74.34 & 66.86 & 65.40 & 86.40 & 68.32 & 49.96 & 45.79 & 39.59 & 48.31 & 58.59 & 61.85 & 77.65 & 52.17 & 62.46 & 31.59 & 51.60 & 47.84\\
 & IoU $>0.5$  & 70.63 & 68.21 & 61.93 & 41.92 & 73.04 & 62.90 & 33.77 & 26.96 & 27.69 & 38.10 & 50.07 & 51.71 & 69.74 & 36.00 & 58.64 & 20.72 & 38.81 & 35.41\\
 & IoU $>0.75$ & 22.26 & 46.34 & 36.98 & 3.50 & 26.59 & 20.47 & 4.13 & 2.48 & 5.53 & 9.39 & 13.21 & 11.25 & 22.61 & 7.37 & 30.53 & 2.70 & 10.07 & 8.69
 \\\hline
 \multirow{3}{*}{\rotatebox{90}{\textbf{S2}}} & IoU $>0.05$ & 80.35 & 88.38 & 66.79 & 47.65 & 83.40 & 71.17 & 63.24 & 46.36 & 71.87 & 29.91 & N/A & 55.36 & 78.02 & 55.17 & 61.55 & 23.19 & 49.30 & 46.64\\
 & IoU $>0.5$ & 67.42 & 85.62 & 62.75 & 26.27 & 65.90 & 59.22 & 44.14 & 30.30 & 56.28 & 24.31 & N/A & 47.00 & 73.82 & 39.49 & 51.56 & 16.95 & 34.95 & 33.11\\
 & IoU $>0.75$  & 18.41 & 60.43 & 33.32 & 2.21 & 6.41 & 14.55 & 4.65 & 1.77 & 12.80 & 7.40 & N/A & 7.54 & 36.94 & 9.45 & 22.1 & 2.46 & 8.68 & 8.05\\ \hline
\end{tabular}
}
\label{tab:results_obj}
\end{table}

\vspace{-3mm}
\subsection{Action Recognition Benchmark}
\label{sec:action_recognition_benchmark}
\noindent\textbf{Challenge:} Given an action segment $A_i = [t_{s_i}, t_{e_i}]$, we aim to classify the segment into its action class, where classes are defined as ${C_a = \{(c_v \in C_V, c_n \in C_N)\}}$, and $c_n$ is the first noun in the narration when multiple nouns are present.
Note that our dataset supports more complex action-level challenges, such as action localisation in the videos of full duration. We decided to focus on the classification challenge first (the segment is provided) since most existing works tackle this challenge.

\begin{table}[t]
\vspace{-1mm}
\begin{center}
\caption{Baseline results for the action recognition challenge}
\label{tab:results_recognition}
\resizebox{\columnwidth}{!}{%
\begin{tabular}{ll|ccc|ccc|ccc|ccc}

                        &        & \multicolumn{3}{c|}{\textbf{Top-1 Accuracy}} & \multicolumn{3}{c|}{\textbf{Top-5 Accuracy}} & \multicolumn{3}{c|}{\textbf{Avg Class Precision}} &\multicolumn{3}{c}{\textbf{Avg Class Recall}} \\\cline{3-14}
                        &        & VERB      & NOUN      & ACTION     & VERB      & NOUN      & ACTION     & VERB   & NOUN  & ACTION & VERB   & NOUN  & ACTION \\\hline
\multirow{6}{*}{\rotatebox{90}{\textbf{S1}}}
                        &Chance/Random &12.62 &1.73 &00.22 &43.39 &08.12 &03.68 &03.67 &01.15 &00.08 &03.67 &01.15 &00.05\\
                        & Largest Class &22.41 &04.50 &01.59 &70.20 &18.89 &14.90 &00.86 &00.06 &00.00 &03.84 &01.40 &00.12\\
                        & 2SCNN (FUSION) & 42.16     & 29.14     & 13.23                & 80.58     & 53.70     & 30.36                & 29.39     & 30.73     & 5.35                      & 14.83     & 21.10     & 04.46 \\
                        & TSN (RGB)      & 45.68     & {\bf36.80}& 19.86                & {\bf85.56}& {\bf64.19}& {\bf41.89}           & {\bf61.64}& 34.32     & 09.96                     & {\bf23.81}& {\bf31.62}& 08.81 \\
                        & TSN (FLOW)     & 42.75     & 17.40     & 09.02                & 79.52     & 39.43     & 21.92                & 21.42     & 13.75     & 02.33                     & 15.58     & 09.51     & 02.06 \\
                        & TSN (FUSION)   & {\bf48.23}& 36.71     & {\bf20.54}           & 84.09     & 62.32     & 39.79                & 47.26     & {\bf35.42}& {\bf10.46}                & 22.33     & 30.53     & {\bf08.83} \\
                        \hline
\multirow{6}{*}{\rotatebox{90}{\textbf{S2}}}
                        & Chance/Random &10.71 &01.89 &00.22 &38.98 &09.31 &03.81 &03.56 &01.08 &00.08 &03.56 &01.08 &00.05\\
                        & Largest Class &22.26 &04.80 &00.10 &63.76 &19.44 &17.17 &00.85 &00.06 &00.00 &03.84 &01.40 &00.12\\
                        & 2SCNN (FUSION) & 36.16     & 18.03     & 07.31                & 71.97     & 38.41     & 19.49                & 18.11     & 15.31     & 02.86                     & 10.52     & 12.55     & 02.69 \\
                        & TSN (RGB)      & 34.89     & 21.82     & 10.11                & {\bf74.56}& 45.34     &{\bf25.33}            & 19.48     & 14.67     & 04.77                     & 11.22     & 17.24     & 05.67 \\
                        & TSN (FLOW)     & {\bf40.08}& 14.51     & 06.73                & 73.40     & 33.77     & 18.64                & 19.98     & 09.48     & 02.08                     & {\bf13.81}& 08.58     & 02.27 \\
                        & TSN (FUSION)   & 39.40     & {\bf22.70}& {\bf10.89}           & 74.29     & {\bf45.72}& 25.26                & {\bf22.54}& {\bf15.33}& {\bf05.60}                & 13.06     & {\bf17.52}&{\bf05.81} \\
\end{tabular}}
\vspace{-1mm}
\caption{Sample baseline action recognition per-class metrics (using TSN fusion)}
\label{tab:results_recognition_top_verbs}
\resizebox{\columnwidth}{!}{%
\begin{tabular}{ll|ccccccccccccccc}
  &            & \multicolumn{15}{c}{\textbf{15 Most Frequent (in Train Set) Verb Classes}} \\\cline{3-17}
  &            & put   & take   & wash   & open   & close   & cut    & mix    & pour   & move   & turn-on & remove  & turn-off & throw & dry  & peel    \\\hline
\multirow{2}{*}{\rotatebox{90}{\textbf{S1}}}
  & RECALL     & 67.51 & 48.27 & 83.19   & 63.32  & 25.45   & 77.64  & 50.20  & 26.32  & 00.00  & 08.28   & 05.11   & 05.45    & 24.18 & 36.49  & 30.43   \\
  & PRECISION  & 36.29 & 43.21 & 63.01   & 69.74  & 75.50   & 68.71  & 68.51  & 60.98  & -      & 46.15   & 53.85   & 66.67    & 75.86 & 81.82  & 51.85   \\
  \hline
\multirow{2}{*}{\rotatebox{90}{\textbf{S2}}}
  & RECALL     & 74.23 & 34.05 & 83.67  & 43.64  & 18.40    & 33.90  & 35.85  & 13.13  & 00.00  & 00.00   & 00.00   & 00.00  & 00.00    & 2.70  & 00.00   \\
  & PRECISION  & 29.60 & 30.68 & 67.06  & 56.28  & 66.67    & 88.89  & 70.37  & 76.47  & -      & -       & 00.00   & -      & -        & 100.0 & 00.00   \\
\end{tabular}}
\end{center}
\vspace{-5mm}
\end{table}

\noindent\textbf{Network Architecture:}
We train the Temporal Segment Network (TSN)~\cite{wang2016tsn} as a state-of-the-art architecture in action recognition, but adjust the output layer to predict both verb and noun classes jointly, with independent losses, as in~\cite{kalogeiton2017}. We use the PyTorch implementation~\cite{tsnpytorch} with the Inception architecture~\cite{szegedy2015going}, batch normalization~\cite{ioffe2015batch} and pre-trained on ImageNet~\cite{imagenet}. 

\noindent\textbf{Implementation Details:}
We train both spatial and temporal streams, the latter on dense optical flow at 30\fps{} extracted using the $\tvl$ algorithm~\cite{zach2007duality} between RGB frames using the formulation $\tvl\left(I_{2t}, I_{2t+3}\right)$ to eliminate optical flicker, and released the computed flow as part of the dataset. We do not perform stratification or weighted sampling, allowing the dataset class imbalance to propagate into the mini-batch.
We train each model on 8 Nvidia P100 GPUs on a single compute node (Nvidia DGX-1) for 80 epochs with a mini-batch size of 512. We set learning rate to 0.01 for spatial and 0.001 for temporal streams decreasing it by a factor of 10 after epochs 20 and 40. After averaging the 25 samples within the action segment each with 10 spatial croppings as in~\cite{wang2016tsn}, we fuse both streams by averaging class predictions with equal weights. All unspecified parameters use the same values as~\cite{wang2016tsn}.

\noindent\textbf{Evaluation Metrics:} We report two sets of metrics: aggregate and per-class, which are equivalent to the class-agnostic and class-aware metrics in~\cite{zhao2017slac}. For aggregate metrics, we compute top-1 and top-5 accuracy for correct predictions of $c_v$, $c_n$ and their combination $(c_v, c_n)$ -- we refer to these as `verb', `noun' and `action'. Accuracy is reported on the full test set. For per-class metrics, we compute precision and recall, for classes with more than 100 samples in training, then average the metrics across classes - these are 26 verb classes, 71 noun classes, and 819 action classes. Per-class metrics for smaller classes are $\approx 0$ as TSN is better suited for classes with sufficient training data.

\noindent\textbf{Results:}
We report results in Table~\ref{tab:results_recognition} for aggregate metrics and per-class metrics. 
We compare TSN (3 segments) to 2SCNN~\cite{simonyan2014two} (1 segment), chance and largest class baselines.
Fused results perform best or are comparable to the best stream (spatial/temporal). The challenge of getting both verb and noun labels correct remains significant for both \textit{seen} (top-1 accuracy 20.5\%) and \textit{unseen} (top-1 accuracy 10.9\%) environments. This implies that for many examples, we only get one of the two labels (verb/noun) right.
Results also show that generalising to \textit{unseen} environments is a harder challenge for actions than it is for objects. 
We give a breakdown per-class metrics for the 15 largest verb classes in Table~\ref{tab:results_recognition_top_verbs}.

\begin{figure}[t!]
\vspace{-2mm}
\centering
\includegraphics[width=\linewidth]{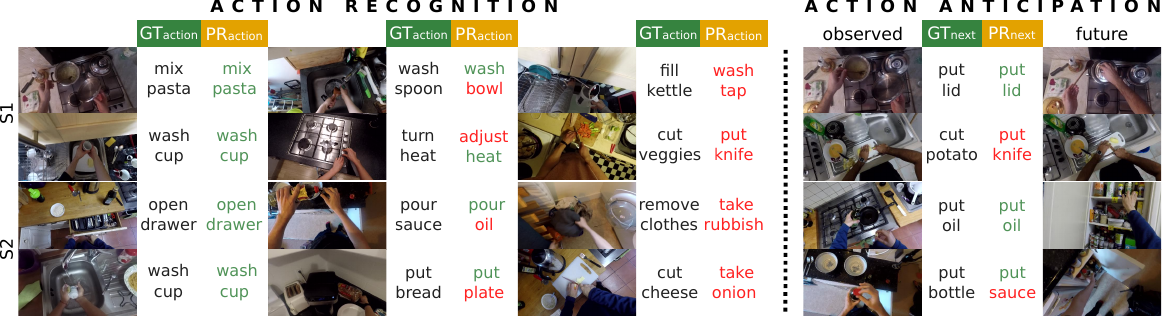}
\vspace{-2mm}
\caption{Qualitative results for the action recognition and anticipation challenges}
\label{fig:recognitionPrediction_qualitative}
\end{figure}

\figurename~\ref{fig:recognitionPrediction_qualitative} reports qualitative results, with success highlighted in green, and failures in red. In the first column both the verb and the noun are correctly predicted, in the second column one of them is correctly predicted, while in the third column both are incorrect. Challenging cases like distinguishing `adjust heat' from turning it on, or pouring soy sauce vs oil are shown.

\vspace{-4mm}
\subsection{Action Anticipation Benchmark}
\vspace{-2mm}
\noindent\textbf{Challenge:} 
Anticipating the next action is a well-mastered skill by humans, and automating it has direct implications in assertive living.
Given any of the upcoming wearable system (e.g. Microsoft Hololens or Google Glass), anticipating the wearer's next action, from a first-person view, could trigger smart home appliances, providing a seamless achievement of the wearer's goals.
Previous works have investigated different anticipation tasks from an egocentric perspective, e.g. predicting future localisation~\cite{park2016egocentric} or next-active object~\cite{furnari2017next}. We here consider the task of forecasting an action before it happens. Let $\tau_a$ be the `anticipation time', how far in advance to recognise the action, and $\tau_o$ be the `observation time', the length of the observed video segment preceding the action. Given an action segment $A_i=[t_{s_i},t_{e_i}]$, we predict
the action class $C_a$ by observing the video segment \emph{preceding} the action start time $t_{s_i}$ by $\tau_a$, that is $[t_{s_i}-(\tau_a+\tau_o),t_{s_i}-\tau_a]$.

\noindent\textbf{Network Architecture:}
As in Sec.~\ref{sec:action_recognition_benchmark}, we train TSN~\cite{wang2016tsn} to provide baseline action anticipation results and compare with 2SCNN~\cite{simonyan2014two}. We feed the model with the video segments preceding annotated actions and train it to predict verb and noun classes jointly as in~\cite{kalogeiton2017}. Similarly to~\cite{vondrick2016anticipating}, we set $\tau_a=1s$. We report results with $\tau_o=1s$, and note that performance drops with longer segments. 

\noindent\textbf{Implementation Details:} Models for both spatial and temporal modalities are trained using a single Nvidia Titan X with a batch size of 64, for $80$ epochs, setting the initial learning rate to $0.001$ and dropping it by a factor of 10 after $30$ and $60$ epochs. Fusion weights spatial and temporal streams with 0.6 and 0.4 respectively. All other parameters use the values specified in~\cite{wang2016tsn}.

\noindent\textbf{Evaluation Metrics:}
We use the same evaluation metrics as in Sec.~\ref{sec:action_recognition_benchmark}.

\noindent\textbf{Results:}
\tablename~\ref{tab:results_anticipation} reports baseline results for the action anticipation challenge. 
As expected, this is a harder challenge than action recognition, and thus we note a drop in performance throughout. Unlike the case of action recognition, the flow stream and fusion do not generally improve performances. TSN often offers small, but consistent improvements over 2SCNN.

\figurename~\ref{fig:recognitionPrediction_qualitative} reports qualitative results. Success examples are highlighted in green, and failure cases in red.
As the qualitative figure shows, the method over-predicts `put' as the next action. Once an object is picked up, the learned model has a tendency to believe it will be put down next. Methods that focus on long-term understanding of the goal, as well as multi-scale history would be needed to circumvent such a tendency. 

\begin{table}[t]
	\centering
	\caption{Baseline results for the action anticipation challenge}
	\label{tab:results_anticipation}
	\resizebox{\columnwidth}{!}{%
		\begin{tabular}{ll|ccc|ccc|ccc|ccc}
			& & \multicolumn{3}{c|}{\textbf{Top-1 Accuracy}} & \multicolumn{3}{c|}{\textbf{Top-5 Accuracy}} & \multicolumn{3}{c|}{\textbf{Avg Class Precision}} &\multicolumn{3}{c}{\textbf{Avg Class Recall}} \\\cline{3-14}
			&  & VERB      & NOUN      & ACTION     & VERB      & NOUN      & ACTION     & VERB   & NOUN  & ACTION & VERB   & NOUN  & ACTION \\\hline
			\multirow{6}{*}{\rotatebox{90}{\textbf{S1}}}   
			
			& 2SCNN (RGB) & 29.76 & 15.15 & 04.32 & 76.03 & 38.56 & 15.21 & 13.76 & 17.19 & 02.48 & 07.32 & 10.72 & 01.81 \\
			& TSN (RGB)      & \textbf{31.81} & \textbf{16.22} & \textbf{06.00} & \textbf{76.56} & \textbf{42.15} & \textbf{18.21} & \textbf{23.91} & 19.13 & \textbf{03.13} & \textbf{09.33} &\textbf{11.93} & \textbf{02.39} \\
			& TSN (FLOW)     & 29.64 & 10.30 & 02.93 & 73.70 & 30.09 & 10.92 & 18.34 & 10.70 & 01.41 & 06.99 & 05.48 & 01.00 \\
			& TSN (FUSION)   & 30.66 & 14.86 & 04.62 & 75.32 & 40.11 & 16.01 & 08.84 & \textbf{21.85} & 02.25 & 06.76 & 09.15 & 01.55 \\
			
			\hline
			\multirow{6}{*}{\rotatebox{90}{\textbf{S2}}} 
			
			& 2SCNN (RGB) & 25.23 & 09.97 & 02.29 & \textbf{68.66} & 27.38 & 09.35 & 16.37 & 06.98 & 00.85 & 05.80 & 06.37 & \textbf{01.14} \\
			& TSN (RGB)  	 & 25.30 & \textbf{10.41} & \textbf{02.39} & 68.32 & \textbf{29.50} & \textbf{09.63} & 07.63 & \textbf{08.79} & 00.80 & 06.06 & \textbf{06.74} & 01.07 \\
			& TSN (FLOW)     & \textbf{25.61} & 08.40 & 01.78 & 67.57 & 24.62 & 08.19 & 10.80 & 04.99 & 01.02 & \textbf{06.34} & 04.72 & 00.84 \\
			& TSN (FUSION)   & 25.37 & 09.76 & 01.74 & 68.25 & 27.24 & 09.05 & \textbf{13.03} & 05.13 & 00.90 & 05.65 & 05.58 & 00.79 \\
			\hline
	\end{tabular}}
\end{table}

\vspace*{-12pt}
\subsubsection{Discussion:} The three defined challenges form the base for higher-level understanding of the wearer's goals. We have shown that existing methods are still far from tackling these tasks with high precision, pointing to exciting future directions. Our dataset lends itself naturally to a variety of less explored tasks. We are planning to provide a wider set of challenges, including action localisation~\cite{Yeung2018}, video parsing~\cite{Sigurdsson2016}, visual dialogue~\cite{visdial}, goal completion~\cite{Heidarivincheh2018} and skill determination~\cite{Hazel2018} (e.g. how good are you at making your eggs for breakfast?). Since real-time performance is crucial in this domain, our leaderboard will reflect this, pressing the community to come up with efficient and effective solutions.

\section{Conclusion and Future Work}
We present the largest and most varied dataset in egocentric vision to date, \EPIC{}, captured in participants' native environments. We collect 55 hours of video data recorded on a head-mounted GoPro, and annotate it with narrations, action segments and object annotations using a pipeline that starts with live commentary of recorded videos by the participants themselves. 
Baseline results on object detection, action recognition and anticipation challenges show the great potential of the dataset for pushing approaches that target fine-grained video understanding to new frontiers.

\section*{Dataset Release:}

\begin{itemize}[leftmargin=*]
\item Dataset sequences, extracted frames and optical flow are available at: \\\textcolor{blue}{\underline{\url{http://dx.doi.org/10.5523/bris.3h91syskeag572hl6tvuovwv4d}}}

\item Annotations, challenge leader-board results and updates and news are available at: \textcolor{blue}{\underline{\url{http://epic-kitchens.github.io}}}
\end{itemize}

  \section*{Acknowledgment}
  
The authors would like to thank all 32 subjects who participated in the dataset collection. 

\noindent The dataset annotation and release has been sponsored by a charitable donation from Nokia Technologies and the University of Bristol's Jean Golding Institute. 

\noindent Research at the University of Bristol is supported by EPSRC DTP, EPSRC GLANCE (EP/N013964/1) and EPSRC LOCATE (EP/N033779/1).

\noindent Research at the University of Catania is sponsored by Piano della Ricerca 2016-2018 – linea di Intervento~2 of DMI. 

\noindent The object detection benchmark baseline results have been helped by code from, and discussions with, Davide Acu{\~{n}}a.


\bibliographystyle{splncs04}

\end{document}